\documentclass{article}

\usepackage[algo2e]{algorithm2e} 
\usepackage{mathtools}
\usepackage{microtype}
\usepackage{graphicx}
\usepackage{subcaption}
\usepackage{booktabs} %
\usepackage{graphicx}
\usepackage[dvipsnames]{xcolor}
\usepackage[inline]{enumitem}
\usepackage{bm}
\usepackage{tikz}
\usepackage{wrapfig}
\usepackage[inline]{enumitem}   
\usepackage{appendix}
\let\appendixpagenameorig\appendixpagename
\renewcommand{\appendixpagename}{\appendixpagenameorig}

\usepackage{hyperref}

\usepackage[accepted]{icml2023}

\usepackage{amsmath}
\usepackage{amssymb}
\usepackage{mathtools}
\usepackage{amsthm}

\usepackage[capitalize,noabbrev]{cleveref}

\theoremstyle{plain}
\newtheorem{theorem}{Theorem}[section]

\theoremstyle{definition}

\theoremstyle{remark}

\usepackage[textsize=tiny]{todonotes}

\icmltitlerunning{GCANs}

\renewcommand{\b}{\mathbf}
\newcommand{\mv}[1]{\b{#1}}  %

\usepackage{amsmath,amsfonts,bm}

\newcommand{\Euc}{\mathrm{E}}
\newcommand{\SEuc}{\mathrm{SE}}
\newcommand{\Pin}{\mathrm{Pin}}
\newcommand{\Spin}{\mathrm{Spin}}
\newcommand{\SO}{\mathrm{SO}}
\renewcommand{\O}{\mathrm{O}}

\def\eqref#1{equation~\ref{#1}}

\def\1{\bm{1}}

\def\rva{{\mathbf{a}}}

\def\rvu{{\mathbf{u}}}
\def\rvv{{\mathbf{v}}}

\def\rvx{{\mathbf{x}}}
\def\rvy{{\mathbf{y}}}
\def\rvz{{\mathbf{z}}}

\DeclareMathAlphabet{\mathsfit}{\encodingdefault}{\sfdefault}{m}{sl}
\SetMathAlphabet{\mathsfit}{bold}{\encodingdefault}{\sfdefault}{bx}{n}

\def\gG{{\mathcal{G}}}

\def\sC{{\mathbb{C}}}

\def\sG{{\mathbb{G}}}

\newcommand{\E}{\mathbb{E}}

\newcommand{\R}{\mathbb{R}}
\newcommand{\G}{\mathbb{G}}

\begin{document}

\twocolumn[
\icmltitle{Geometric Clifford Algebra Networks}

\begin{icmlauthorlist}
\icmlauthor{David Ruhe}{msrint}
\icmlauthor{Jayesh~K.~Gupta}{msrj}
\icmlauthor{Steven de Keninck}{uvas}
\icmlauthor{Max Welling}{msrai}
\icmlauthor{Johannes Brandstetter}{msrai}
\end{icmlauthorlist}

\icmlaffiliation{msrint}{Work done during internship at Microsoft Research.}
\icmlaffiliation{uvas}{University of Amsterdam}
\icmlaffiliation{msrj}{Microsoft Autonomous Systems and Robotics Research}
\icmlaffiliation{msrai}{Microsoft Research AI4Science}

\icmlcorrespondingauthor{David Ruhe}{david.ruhe@gmail.com}
\icmlcorrespondingauthor{Johannes Brandstetter}{johannesb@microsoft.com}

\icmlkeywords{Machine Learning, ICML}

\vskip 0.3in
]

\printAffiliationsAndNotice{}  %

\begin{abstract}
    We propose Geometric Clifford Algebra Networks (GCANs) for modeling dynamical systems. 
    GCANs are based on symmetry group transformations using geometric (Clifford) algebras. 
    We first review the quintessence of modern (plane-based) geometric algebra, which builds on isometries encoded as elements of the $\Pin(p,q,r)$ group.
    We then propose the concept of group action layers, which linearly combine object transformations using pre-specified group actions. 
    Together with a new activation and normalization scheme, these layers serve as adjustable \emph{geometric templates}
    that can be refined via gradient descent.
    Theoretical advantages are strongly reflected in the modeling of three-dimensional rigid body transformations as well as large-scale fluid dynamics simulations, showing significantly improved performance over traditional methods.
\end{abstract}

\section{Introduction}\label{sec:intro}

Equipping neural networks with geometric priors has led to many recent successes.
For instance, in group equivariant deep learning \citep{cohen2016group, weiler20183d, bronstein2021geometric, weiler2021coordinate}, neural networks are constructed to be equivariant or invariant to group actions applied to the input data.

In this work, 
we focus on tasks where we expect that the target function is a \emph{geometric transformation} of the input data.
Such functions arise ubiquitously in the science of dynamical systems, which is the core experimental domain of this work. 
\begin{figure}[!htb]
\centering
\includegraphics[width=0.99\linewidth]{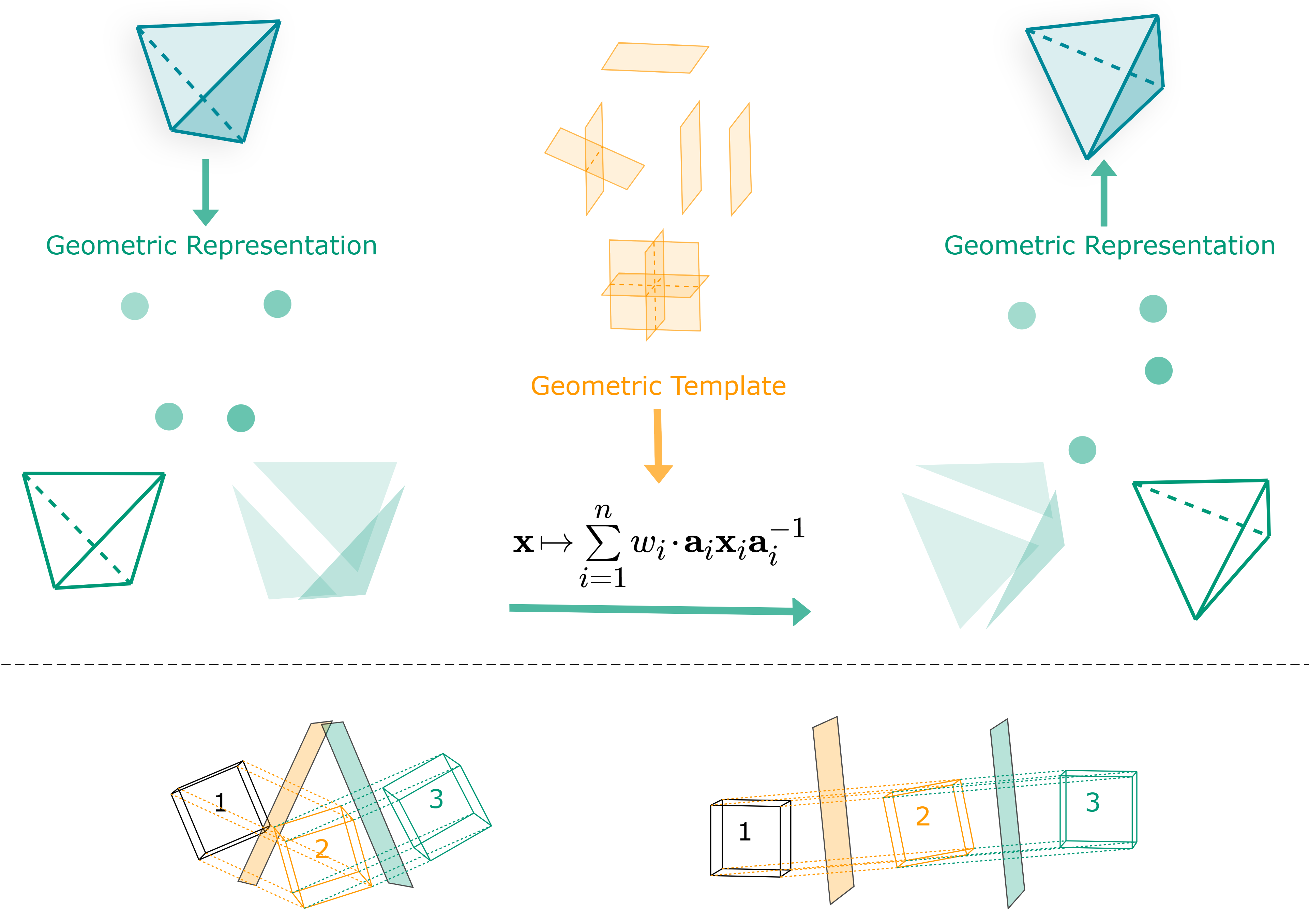}

\caption{GCANs: Geometric algebra allows us to express data 
as objects which can simultaneously be interpreted as group elements (top). GCANs parameterize linear combinations of learnable group actions, 
ensuring that even randomly initialized models form a composition of geometric transformations. GCANs are thus adjustable \emph{geometric templates}. 
Bottom: geometric transformations can be constructed from compositions of reflections. Two reflections in intersecting planes yield a rotation (left), and two reflections in parallel planes yield a translation (right).}
\label{fig:GA-NNs}
\end{figure}
Neural surrogates for solving these schemes have been proposed in fluid dynamics \citep{li2020fourier, kochkov2021machine, lu2021learning, 
rasp2021data,
pathak2022fourcastnet,
bi2022pangu, 
lam2022graphcast,nguyen2023climax},
molecular dynamics \citep{mardt2018vampnets, zhong2019symplectic, greydanus2019hamiltonian, mattheakis2019physical, li2020neural}, or astrophysics \citep{tamayo2016machine, cranmer2021bayesian}.
Typically, the objective is to predict with high precision how a system will evolve based on various initial conditions.
This is a challenging task, given that the underlying dynamics can be highly unstable or chaotic.

We introduce Geometric Clifford Algebra Networks (GCANs)
as a new approach to incorporate geometry-guided transformations into neural networks using \emph{geometric algebra}.
Geometric algebra~\citep{CliffordSketch, hestenes1966STA, dorst2002geometric,mann2002geometric, dorst2010geometric, artin2016geometric}
is an algebraic framework based on real Clifford algebras that is particularly well suited to handle and model computational geometry. 
It has several intriguing properties and advantages over other frameworks, such as classical linear algebra.
For example, it naturally and efficiently encodes the transformations and the invariant elements of classic geometries.    
Additionally, in geometric algebra, objects transform covariantly with transformations of space. 
This means that a single function can transform multiple types of objects, including vectors, points, lines, and planes. 
Finally, geometric algebra generalizes over dimensions in the sense that transformations and objects are constructed consistently regardless of the dimensionality of the space.

GCANs are built around the concept of \emph{group action layers}, which linearly combine pre-specified group actions to form neural representations.
This process can be efficiently implemented using geometric algebra, which encodes both objects and transformations elegantly.
By exploiting knowledge of the transformation group actions that govern the underlying dynamics, our approach provides a geometric prior complementary to the symmetry and scale separation principles discussed in \citet{bronstein2021geometric}.
In order to ensure the preservation of the vector space of the input representations, we further introduce a specific type of nonlinearity and a new form of normalization.
The resulting network can be seen as a composition of geometric transformations, with an uninitialized network serving as a \emph{geometric template} that can be refined through gradient descent learning.

Recently, \emph{Clifford neural layers} \citep{brandstetter2022clifford}, which can encode spatial primitives such as scalars and vectors 
into single entities, were proposed.
Most notably, Clifford neural networks can already be seen as geometric templates by allowing Fourier transforms on \emph{multivectors} and geometric products beyond complex numbers or quaternions.
However, most Clifford neural layers lack certain geometric guarantees.
For example, vector-valued input features 
might result in multivector-valued quantities that are difficult to interpret geometrically. 
We will see that specific layers, such as the rotational Clifford convolution layers, are already close to a specific instance of GCAN layers proposed in this work. In fact, the compelling performance of these layers
served as motivation for this work.

The theoretical advantages of GCANs are reflected in various dynamical systems modeling tasks.
The strong inductive bias of GCANs enhances generalization in low-data regimes and allows for more efficient optimization when there is plenty of data and, therefore, outperforms baselines in both these regimes.
We demonstrate these advantages on a rigid body transformation task, where we simulate the motion of Tetris objects in free space.
Next, we show excellent performance on two large-scale PDE modeling tasks of fluid dynamic problems, i.e., weather prediction based on the shallow water equations and fluid systems described by incompressible Navier-Stokes equations. 

\section{Geometric algebra}
This section presents a formalization of Euclidean geometry from a geometric algebra perspective, primarily based on \citet{roelfs2021graded} and \citet{de2022normalization}.
We derive how the $\Pin(n)$ group can model isometries (distance-preserving transformations of metric spaces) and how 
to use it to obtain geometric templates.

\textbf{The Pin$(n)$ group. }
We start the formalization by constructing symmetries (isometries) using \emph{reflections} as our foundation. 
A reflection is a map from a Euclidean space to itself with a hyperplane as a set of fixed points in which the space gets mirrored.
The mathematician Hamilton observed that the composition of two reflections through intersecting planes results in a rotation.
This idea is presented in \Cref{fig:GA-NNs} (bottom) and can be generalized to the following theorem.
\begin{theorem}
\label{thm:cartan}
Cartan-Dieudonn\'e. Every orthogonal transformation of an $n$-dimensional space can be decomposed into, at most, $n$ reflections in hyperplanes.
\end{theorem}
It is worth noting that isometries composed of an odd number of reflections change the chirality (handedness) of the space, which is often an unwanted property.
We refer to such isometries as \emph{improper}.
In an $n$-dimensional space, compositions of reflections construct the $\Pin(n)$ group. 
A group $(\gG, \circ)$ is 
a non-empty set $\gG$ equipped with a binary composition operator $\circ: \gG \times \gG \rightarrow \gG$ (written here with juxtaposition) that satisfies 
\begin{enumerate*}[label=(\roman*)]
    \item  closure, i.e., for $u, v \in \gG: uv \in \gG$,
    \item associativity, i.e., for $u,v,w \in \gG: (uv)w = u(vw)$,
    \item identity, i.e., there is an element $1 \in \gG$, such that for $u \in \gG: 1u = u = u1$, and
    \item inverse, i.e., for $u \in \gG$ there is an inverse element $u^{-1} \in \gG$ such that $uu^{-1} = 1 = u^{-1}u$.
\end{enumerate*}
Since compositions of reflections (like the ones shown in \Cref{fig:GA-NNs}) satisfy all these conditions, they form a group. 
That is, any element $u \in \Pin(n)$ can be written as composition of $k$ linearly independent reflections:
$
    u=u_1 \dots u_k.
$

\textbf{Group action: conjugation. }
A group action on a space is a group homomorphism of the group into the group of transformations of that space.
A group can act on itself by the \emph{conjugation} rule, which is a specific map $\gG \times \gG \to \gG$.
For $u,v \in \Pin(n)$, we let $u$ act on $v$ via 
\begin{align}
u[v] \mapsto uvu^{-1} \ , \label{eq:group_conj}
\end{align}
where the group composition is used twice and $uvu^{-1}~\in~\Pin(n)$.
This \emph{sandwich product} tells us how we let a group element act on another (e.g., reflecting a reflection).
Intuitively, it mimics what we would do when asked, for example, to write our name upside down: we first rotate the page, write, and rotate the page back.
In the following sections, we will see that geometric algebra forms a framework where the objects (e.g., vectors) we want to act on can be interpreted as group elements and vice versa.
This allows us to also apply the conjugation rule not only to group elements but also to geometric primitives.

\textbf{Geometric algebra. }
Equipped with an understanding of how compositions of reflections yield higher-order orthogonal transformations and how we can let group elements act, we can construct an algebraic implementation of these ideas~\citep{roelfs2021graded}
using geometric algebra.
Geometric algebra is an emerging tool 
to model computational geometry~\citep{dorst2010geometric}
based on real Clifford algebras.
The geometric product of the algebra allows for intuitive expressions of geometric transformations, making this a natural choice.
In an $n$-dimensional geometric algebra\footnote{Technically, there are no differences between geometric and (real) Clifford algebra. In fact, Clifford himself chose ``geometric algebra''. However, it is common practice to use Clifford algebra when primarily interested in mathematical concerns and geometric algebra when interested in geometry. 
} $\G_{pqr}$\footnote{$\G_{pqr}$ corresponds to $Cl_{p, q, r}(\R)$ when using the notation of~\citet{brandstetter2022clifford}.} of $n=p+q+r$, we choose $p$ positive, $q$ negative, and $r$ null basis vectors $e_i$ 
with
\begin{align}
    & e_ie_i \in \left \{+1,-1,0 \right \}, \quad e_i e_j = -e_j e_i \quad (i \neq j) \ , \label{eq:quadratic_antisymmetric}
\end{align}
where the juxtaposition $e_i e_j$ denotes the algebra's bilinear product.
For example, the simplest three-dimensional algebra is $\sG_{3,0,0}$, 
with $p=3$, $q=0$, and $r=0$.
A product of $k$ basis vectors is a basis $k$-blade where the \emph{grade} of the blade is the dimension of the subspace it represents.
In this way, vectors with basis components $e_i$ are $1$-blades.
$2$-blades are of the form $e_i e_j$, and so on.
In general, an $n$-dimensional vector space yields $2^n$ basis blades.
The highest grade basis blade $I := e_{1}\ldots e_{n}$ is also known as the \emph{pseudoscalar}.
We speak of $k$\emph{-vectors} by homogeneously combining basis blades of grade $k$. 
Vectors, like in linear algebra, are linear combinations of $1$-blades.
\emph{Bivectors} are formed from $2$-blades, etc.
A \emph{multivector} $\mv x \in \sG_{p, q, r}$ is a sum of $k$-vectors, i.e., $\mv x = [\mv x]_0 + [\mv x]_1 + [\mv x]_2 + \ldots + [\mv x]_n$, where $[\mv x]_k$ denotes the $k$-vector part of $\mv x$.  
As an example, $\G_{3,0,0}$ has $2^3=8$ basis blades, 
where a multivector $\rvx$ is represented via
\begin{equation}
    \begin{aligned}
        \mv x = &\underbrace{x_0 1}_{\text{Scalar}} + \underbrace{x_1 e_1 + x_2 e_2 + x_3 e_3}_{\text{Vector}}  \\ &+ \underbrace{x_{12}e_{12} + x_{13}e_{13} + x_{23}e_{23}}_{\text{Bivector}} + \underbrace{x_{123}e_{123}}_{\text{Trivector}} \ .
    \end{aligned}
\end{equation}
Here, we used $e_{ij} :=e_i e_j$.
It is worth mentioning that the set of basis blades is closed under multiplication with its elements using \Cref{eq:quadratic_antisymmetric}.
The specific choices of algebra (determined by $p, q$, and $r$) allows for efficient modeling of many types of geometry.

\textbf{Geometric product. }
Multiplication in the algebra is realized via the \emph{geometric product}: a bilinear operation between multivectors.
For arbitrary multivectors $\rvx$, $\rvy$, $\rvz \in \G_{pqr}$, and scalar $\lambda$, the geometric product has the following properties:
\begin{enumerate*}[label=(\roman*)]
    \item  closure, i.e., $\rvx\rvy \in \G_{pqr}$,
    \item associativity, i.e., $(\rvx\rvy)\rvz = \rvx(\rvy\rvz)$,
    \item commutative scalar multiplication, i.e., $\lambda\rvx = \rvx\lambda$, 
    \item distributivity over addition, i.e., $ \rvx(\rvy+\rvz) = \rvx\rvy + \rvx\rvz$, and
    \item Vectors square to scalars given by a metric norm.
\end{enumerate*}
The geometric product is in general non-commutative: $\rvx\rvy \neq \rvy\rvx$.
It can also be applied to (combinations of) lower-grade elements.
For example, as shown in \Cref{supp:ga}, 
for vectors ($1$-vectors) it exactly results in an inner product part and an antisymmetric part associated with a bivector. 
In this case, the geometric product directly measures their
similarities as well as their differences.

\textbf{Representing elements of Pin$(p,q,r)$. } 
We can use the geometric algebra $\G_{pqr}$ to represent elements of $\Pin(p,q,r)$, where $p+q+r=n$ for an $n$-dimensional space with $p$ positive, $q$ negative, and $r$ zero dimensions \footnote{For further theory on representing groups in the geometric algebra, consider dr{cite}}.
We saw that the fundamental isometry (from which we build $\Pin(p, q, r)$) is a reflection.
\begin{figure}
    \centering
    \includegraphics[width=\linewidth]{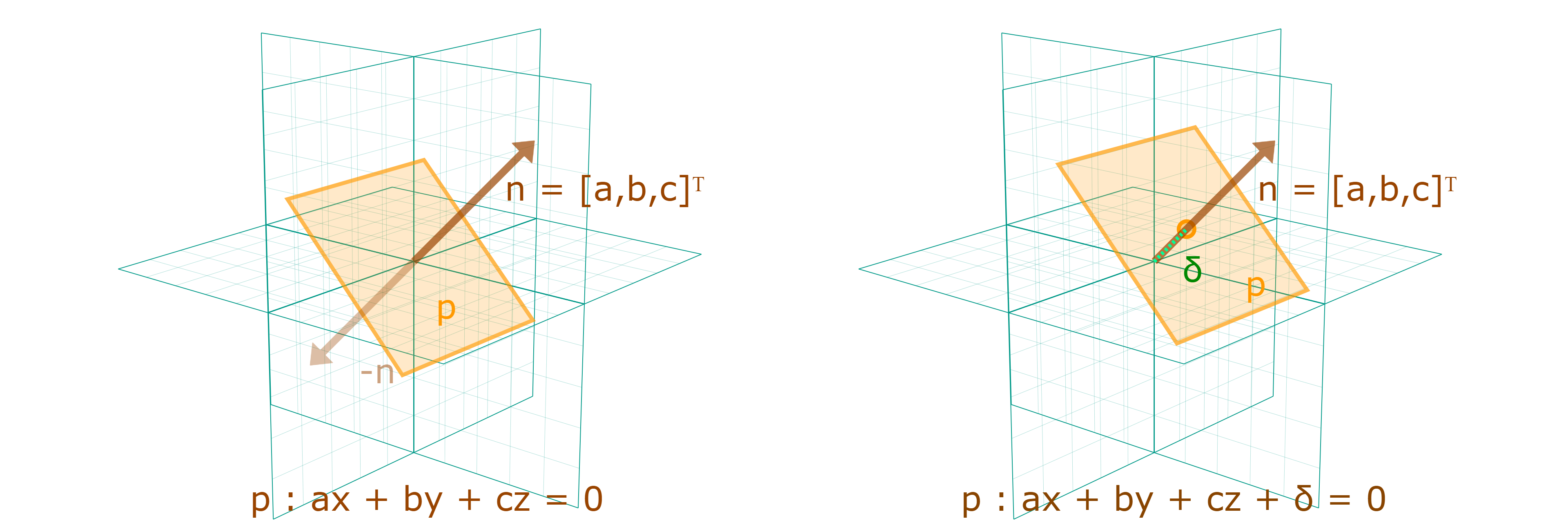}
    \caption{Left, a plane through the origin identified by a normal vector $n = \begin{bmatrix} a & b & c \end{bmatrix}^T$, or equivalently a linear equation $p: ax + by + cz = 0$. %
    Right, a general plane represented by a normal vector $n$ and distance $\delta$, or equivalently by the linear equation $p: ax + by + cz + \delta = 0.$}
    \label{fig:vectors_planes}
\end{figure}
\begin{figure*}
    \centering
    \includegraphics[width=1.0\linewidth]{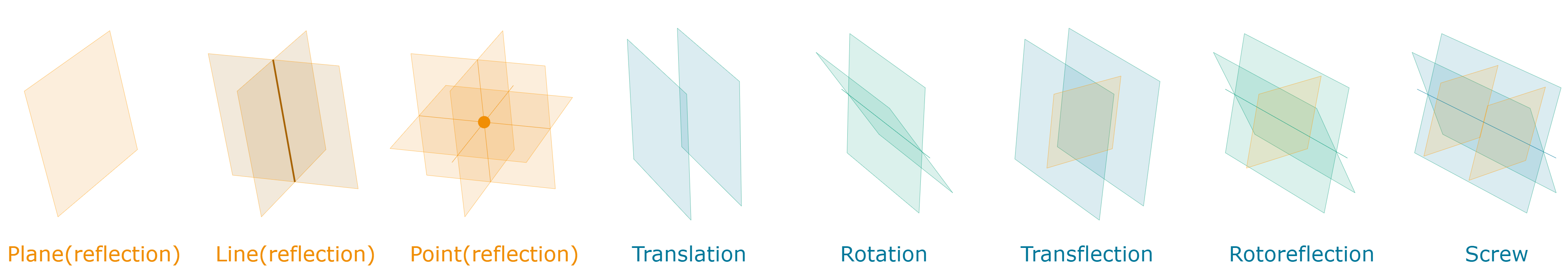}
    \caption{All elements of the Euclidean group can be represented as compositions of reflections in planes. The orange $k$-blades are compositions of orthogonal planes, and represent simultaneously the points, lines, planes as well as the reflections in these elements. In green: compositions of reflections in arbitrary planes make up all isometric transformations.    
    }
    \label{fig:compose-plane-ref}
\end{figure*}
To identify reflections, we use the fact that a hyperplane through the origin $p: ax + by + cz + \ldots = 0$ can be mapped onto grade $1$ elements of $\G_{pqr}$ via
\begin{align}
     ax + by + cz + \ldots = 0 \leftrightarrow \mv u : = ae_1 + be_2 + ce_3 + \dots \ .
     \label{eq:map_planes_vectors}
\end{align}
The grade 1 elements form a vector perpendicular to the surface, i.e., $n:=[a, b, c]^\top$ and $[\mv u]_1 = n$.
Note that $\rvu \in \G_{pqr}$, but the $(k \neq 1)$-blades are simply left zero.
This identification is illustrated for the three-dimensional case in \Cref{fig:vectors_planes}.
Note that, using \Cref{eq:map_planes_vectors}, two unit normals lead to the same geometric plane: one being the negation of the other (also displayed in \Cref{fig:vectors_planes}).
As such, normalized vectors map 2-to-1 to planes through the origin. 
Now that we have an algebraic implementation of a hyperplane, 
it can be shown (\Cref{supp:ga}) that a reflection through that plane amounts to (using geometric products) 
\begin{align}
    \mv v \mapsto - \mv u \mv v \mv u^{-1} \ , \label{eq:reflection_ga}
\end{align}
where $\mv u, \mv v \in \sG_{p, q, r}$ are vectors ($1$-vectors in the algebra) and $\mv u^{-1}$ is the multiplicative inverse such that $\mv u \mv u^{-1}=1$.
The minus sign of \Cref{eq:reflection_ga} comes from the fact that we use two $1$-vectors, as explained below.
We next use the fact that any $\Pin(p, q, r)$ group element can be written as a composition of $k$ linearly independent reflections.
Composition is now rather straightforward: we take \Cref{eq:reflection_ga} and apply the sandwich structure again using geometric products:
\begin{align}
    \mv v \mapsto \mv u_2 \mv u_1 \, \mv v \, \mv u_1^{-1} \mv u_2^{-1} = \left(\mv u_2 \mv u_1 \right) \mv v \left (\mv u_2 \mv u_1 \right)^{-1} \, , \label{eq:bireflection}
\end{align}
creating a bireflection.
We see how we can use the associativity of the geometric product to compose reflections.

As such, we henceforth treat elements of $\Pin(p, q, r)$ as compositions of $1$-vectors in the algebra, as opposed to abstract compositions of reflections. 
In general, for two group elements $\mv u :=\mv u_1 \mv u_2 \dots \mv u_k \in \Pin(p, q, r)$ and $\mv v:=\mv v_1 \mv v_2 \dots \mv v_l \in \Pin(p,q,r)$,
where $\rvu$ is a $k$-reflection and $\rvv$ is an $l$-reflection, the group action of $\rvu$ on $\rvv$ is:
\begin{align}
    \mv v \mapsto \mv u[\mv v] := (-1)^{kl} \mv u \mv v \mv u^{-1} \ , \label{eq:pin_ga_group_action}
\end{align}
obtaining a similar conjugation rule to \Cref{eq:group_conj}.
The prefactor ensures that we obtain the correct orientation of space.
For example, when $\mv u$ is a reflection ($k=1$), $\mv u[\mv u]=-\mv u$,
meaning that reflecting $\mv u$ in itself reverses its orientation. 
Taking the geometric product $\mv u_1 \mv u_2$ of two vectors (each parameterizing a reflection) with $\mv u_1, \mv u_2 \in \sG_{p, q, r}$ yields bivector components.
Considering Hamilton's observation, bivectors thus parameterize rotations.
Composing more reflections yields higher-order blades, parameterizing higher-order isometries.

\begin{table*}
\footnotesize
\centering

\begin{tabular}{@{}llll@{}} 
\toprule
Reflections & Group element [$\O(3)$]& Invariant subspace$^\dagger$ & Algebra element [$\sG_{3,0,0}$]   \\ \midrule
$0$                   & Identity              & Volume             & Scalar    \\
$1$                   & Reflection              & Plane              & Vector    \\
$2$                   & Rotation                & Line              & Bivector  \\
$3$                   & Rotoreflection          & Point             & Trivector \\ 
\multicolumn{4}{l}{\begin{scriptsize}$^\dagger$These subspaces all pass through the origin.\end{scriptsize}}
\\ \bottomrule
\end{tabular}

\vspace{1em}

\begin{tabular}{@{}llll@{}} 
\toprule
Reflections & Group element [$\Euc(3)$]& Invariant subspace & Algebra element [$\sG_{3,0,1}$]   \\ \midrule
$0$                   & Identity              & Volume             & Scalar    \\
$1$                   & Reflection              & Plane             & Vector    \\
$2$                   & Rotation/translation    & Line              & Bivector  \\
$3$                   & Roto/transflection      & Point             & Trivector \\ 
$4$                   & Screw                   & Origin            & Quadvector \\ \bottomrule
\end{tabular}
\vspace{1.3em}
    \caption{Overview of elements of $\Pin(3, 0, 0)$ (top), and $\Pin(3, 0, 1)$ (bottom). This table relates their group elements, i.e., composition of reflections, to O$(3)$ and E$(3)$ group elements, spatial primitives (identified with invariant subspaces of the transformations), and how they are encoded in geometric algebra. 
    Note that $\sG_{3,0,1}$ allows us to encode translations.}
    \label{tab:ga_overview}
\end{table*}

Note that $\mv u$ and $-\mv u$ parameterize the same isometry since the sign gets canceled in the sandwich structure of \Cref{eq:pin_ga_group_action}.
This is visualized in \Cref{fig:vectors_planes}, where two vectors parameterize the same plane used to reflect in.
We therefore have a $2$-to-$1$ map to any orthogonal transformation, which makes $\Pin(n)$ the double cover of $\O(n)$.
I.e., each group element in $\O(n)$ identifies with two elements in $\Pin(n)$. 
By excluding improper isometries, we obtain the $\Spin(n)$ group, the double cover of the special orthogonal group $\SO(n)$, which is the group of $n$-dimensional orthogonal transformations excluding reflections.

\textbf{Projective geometric algebra. }
We now take a closer look at an instantiation of $\Pin(p,q,r)$: the projective geometric algebra $\sG_{3, 0, 1}$, which is well-suited to model transformations in three-dimensional space.
Note that the dimensionality of the algebra is higher than that of the physical space.
This is a recurring theme in geometric algebra: higher-dimensional algebras are used to \emph{model} the underlying space.
For example, $e_0^2=\{-1, 0, 1 \}$ leads to hyperbolic, projective, and Euclidean geometry, respectively, where we call the fourth `special' basis vector $e_0$.
The inclusion of the zero element $e_0^2 = 0$ in the algebra allows us to obtain planes that do not pass through the origin of the physical space,
which resembles the extra dimension that we are acquainted to when using homogeneous coordinate systems:
\begin{align}
    \hspace{-0.1343cm}
    ax + by + cz + \delta = 0 \leftrightarrow \mv u = ae_1 + be_2 + ce_3 + \delta e_0
\end{align}
This identification is in \Cref{fig:vectors_planes}.
Consequently, we are free to construct two \emph{parallel} planes, which, as depicted in \Cref{fig:GA-NNs}, can be used to translate.
Two intersecting planes still create rotations, but now around a line not necessarily through the origin.
Three reflections are improper rotations and reflections, and four reflections lead to screw motions.
By including translations,
$\Pin(3, 0, 1)$ is the double cover of the three-dimensional Euclidean group $\Euc(3)=\O(3)\rtimes \R^3$,
which is the semi-direct product of $\O(3)$ and the translation group $\R^3$.
$\Euc(3)$ contains all the transformations of
three-dimensional Euclidean space that preserve the Euclidean distance between any two points, i.e., translations, rotations, and reflections.
As such, we can work with all the rigid motions of Euclidean space by composing reflections (\Cref{fig:compose-plane-ref}).
Similarly, $\Spin(3, 0, 1)$, which excludes improper isometries, i.e., those composed of an odd number of reflections, is the double cover of the special Euclidean group $\mathrm{SE}(3)=\SO(3)\rtimes \R^3$,
the group of three-dimensional Euclidean isometries excluding reflections.

\textbf{Representing data as elements of \textbf{Pin}$(p,q,r)$}. 
We discussed how composing reflections allows us to construct group actions of, e.g., $\Euc(3)$.
However, instead of acting on group elements, we are in practice interested in acting on objects such as vectors, planes, lines, or points.
We can naturally construct these by identifying the \emph{invariant subspaces} (symmetry elements) of the group elements.
This was already shown for reflections: they were constructed from planes, which used the $1$-vector components of the algebra.
Thus, we can relate $1$-vectors, planes, and reflections with each other and, most notably, as elements of $\Pin(p,q,r)$.
Similarly, a bireflection (\Cref{eq:bireflection}) computes a rotation, which preserves a line (see, e.g., \Cref{fig:compose-plane-ref}).
Computing the element $\mv u_1 \mv u_2 \in \Pin(3, 0, 1)$ using the geometric product yields a bivector.
Bivectors ($2$-vectors) thus parameterize lines, i.e., if our data represents a line, we can use the bivector components 
of the respective algebra to represent it, and transform it using \Cref{eq:pin_ga_group_action}.
In this way, we can determine the transformation of any spatial object using conjugation. 
The group action is now not on $G$, but rather on $X$. 
That is, we consider mappings of the form
$\gG \times X \to X$ instead of $\gG \times \gG \to \gG$, where $X$ is the space we are interested in.
Summarizing, the relationship between $\Pin(p, q, r)$ group elements, spatial primitives, and algebraic elements is shown in \Cref{tab:ga_overview}.
Specifically, we can identify each row's elements with each other.

\section{Geometric algebra neural networks}
In the following, we describe the building blocks of GCANs.
Crucially, we want to ensure two properties: 
\begin{enumerate*}[label=(\roman*)]
    \item inputs always map back to their source vector space, and
    \item inputs are transformed by linear combinations of group actions.
\end{enumerate*}
We call such neural networks \emph{geometric templates}.
For CGANs specifically, \begin{enumerate*}[label=(\roman*)]\item means the input grades are unchanged or, in other words, $k$-vectors map to $k$-vectors, and 
\item means that we use linear combinations of $\Pin(p, q, r)$ actions.
\end{enumerate*}

\textbf{Group action layers. }
We start by introducing the general concept of a group action layer.
Let $\gG$ be a group, $X$ a vector space, $\alpha: \gG \times X  \to X$ a group action, and
$c$ a number of input channels.
A group action layer then in general form amounts to
\begin{align}
    x \mapsto T_{g, w}(x) := \sum_{i=1}^c w_i \cdot \alpha(g_i, x_i) \ ,
\end{align}
where we put $g:=(g_i)_{i=1,\dots,c}$
with $g_i \in G$,
$x:=(x_i)_{i=1,\dots,c}$ with $x_i \in X$,
and $w:=(w_i)_{i=1,\dots,c}$ with $w_i \in \mathbb{R}$.
The \emph{scalar} parameters $w$
determine how the actions are linearly combined (``$\cdot$'' denotes scalar multiplication).
This linear combination of group actions ensures that the source object transforms in a geometrically consistent way, regardless of whether it is a point, line, vector, sphere, etc.
Both linear and convolutional neural network layers can be constructed from this general notion.
As an example, let $X:=\mathbb{R}$ and $\gG:=\mathrm{Aff}(1)$ be the one-dimensional affine group of scaling and addition.
Then $g_i=(a_i, b_i) \in \gG$ and $\alpha(g_i, x_i)=a_i x_i + b_i$.
In this case, we recover the standard linear layers, where
\begin{align}
    T_{g, w}(x) = \sum_{i=1}^c w_i \cdot \left(a_i x_i + b_i\right) = \sum_{i=1}^c w'_i x_i + b' \ , \label{eq:scalar_linear_layer}
\end{align}
with $w'_i:=w_i a_i$ and $b':=\sum_{i=1}^c w_i b_i$.

\textbf{GCA linear layers. }
Using geometric algebra, we can now easily encode group transformations and objects into group action layers.
In this case,
\begin{align}
    T_{g, w}(\mv x)=\sum_{i=1}^c w_i \cdot  \mv a_i \mv x_i \mv a_i^{-1} \,, \label{eq:ga_linear}
\end{align}
where now $\rvx_i \in X:=\sG_{p, q, r}$, and $\rva_i \in \gG:=\Spin(p, q, r)$.
The sandwich operation, determined by $\mv a_i$, represents the group action and can be parameterized.
For example, to encode rotations, we parameterize bivector components and optimize them using techniques like stochastic gradient descent.
This can be viewed as an extension of Clifford neural layers~\citep{brandstetter2022clifford} where the geometric product is replaced with the sandwich product, and the transformations are linearly mixed via the scalar parameters $w$.
\Cref{eq:ga_linear} ensures that when $\rvx_i$ is a sum of $k$-vectors, each $k$-vector transforms as a $k$-vector, i.e., object types are preserved (see \Cref{thm:outermorphism}).
For example, when $\mv x_i$ are $1$-vectors, $T_{g, w}(\rvx)$ will be a $1$-vector.
Clifford~\citep{pearson2003clifford}, complex~\citep{Trabelsi2018DeepCN}, or quaternion~\citep{parcollet2020survey} networks generally do not exhibit this property, meaning that a $k$-vector would transform into an unstructured multivector\footnote{The complex and Hamilton (quaternion) product both are instances of Clifford's geometric product.}.
The flexibility of geometric algebra allows the practitioner to determine which group actions to parameterize and how to efficiently represent the data.
Furthermore, due to the sparsity of the sandwich product, GCA layers scale better with the algebra dimension than Clifford layers.
\begin{figure}
\includegraphics[width=\linewidth]{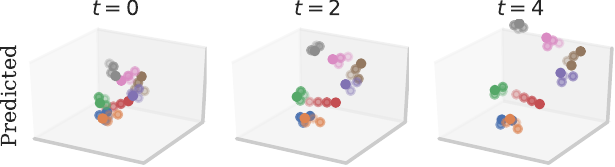} \\
\includegraphics[width=\linewidth]{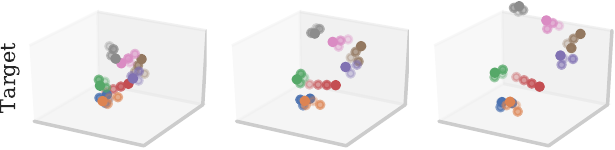}
\caption{Tetris trajectories. Exemplary predicted (top) and ground-truth (bottom) states.
Predictions are obtained by the GCA-GNN model when using 16384 training trajectories.}
\label{fig:tetris_rollout}
\end{figure}

Note that \Cref{eq:ga_linear} computes a ``single-channel'' transformation, similar to how \Cref{eq:scalar_linear_layer} computes a single-channel value (typically referred to as a neuron).
In practice, we extend this to a specified number of output channels by applying the linear transformations in parallel using different weights and transformations.

\textbf{GCA nonlinearity and normalization. }
We are interested in nonlinearities that support the idea of geometric templates. 
Therefore, we propose the following \emph{Multivector Sigmoid Linear Unit} that gates \citep{ramachandran2017searching, weiler20183d, sabour2017dynamic} a $k$-vector by applying
\begin{align}
     [\mv x]_k \mapsto \mathrm{MSiLU}_k\left(\mv x\right) := \sigma \left(f_k(\mv x)\right) \cdot   [\mv x]_k\,,
\end{align}
where $f_k(\mv x): \sG_{p, q, r} \to \mathbb{R}$ and $\sigma$ is the logistic function.
We choose this definition to ensure the geometric template: by only scaling, a $k$-vector remains a $k$-vector.
In this work, we restrict $f_k$ to be a linear combination of the multivector components, i.e.,
$f_k(\mv x) := \sum_{i=1}^{2^n} \beta_{k, i} \cdot \rvx_i$,
where $\rvx_{i}$ denotes the $i$-th blade component of $\mv x$.
Also, $\beta_{k, i}$ are either free scalar parameters or fixed to $\beta_{k, i} := 1$ or $\beta_{k, i}:= 1 / m$, resulting in summation or averaging, respectively.
Similarly, we normalize a $k$-vector by applying a modified version of \emph{group normalization}~\citep{wu2018group}, where
\begin{align}
    [\rvx]_k \mapsto s_k \frac{[\rvx]_k - \E[[\rvx]_k]}{\E[\lVert [\rvx]_k \rVert]} \ .
\end{align}
The empirical average of $[\rvx]_k$ is computed over the number of channels specified by the group size hyperparameter, and rescaled using a learnable scalar $s_k$ through 
$
    [\rvx]_k \mapsto s_k \cdot [\rvx]_k\ ,
$
which again only scales the $k$-vector.

\section{Experiments}
\begin{figure*}[!htb]
    \centering
    \includegraphics[width=\linewidth]{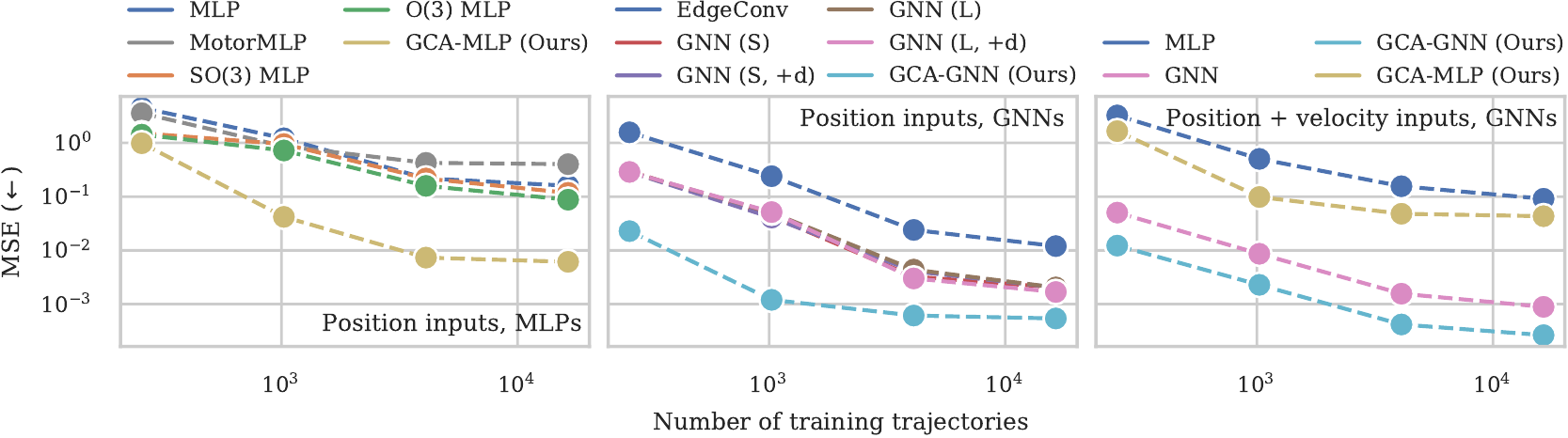}
    \caption{
    Test MSE results of the Tetris experiment as a function of the number of training trajectories. Left: comparison of different MLP models, center: comparison of different GNN models, right: comparison of the best MLP and GNN models when velocities are included.
    }
    \label{fig:tetris_figure}
\end{figure*}
\subsection{Tetris}
This experiment shows the ability of GCANs to model complex object trajectories.
We subject Tetris objects~\citep{thomas2018tensor}, which are initially located at the origin, to random translations and rotations around their respective centers of mass.
The rotations and translations are sampled conditionally, introducing a correlation between the objects.
We further apply conditional Gaussian noise to the individual parts of each object. 
The objects move outward from the origin in an exploding fashion, continuously rotating around their own centers of mass.
Given four input time steps, the model's objective is to accurately predict the following four time steps.
To do so, it has to infer the positions, velocities, rotation axes, and angular velocities and apply them to future time steps, see \Cref{fig:tetris_rollout}).

We use $\sG_{3, 0, 1}$, as it is particularly well-suited to model Euclidean rigid motions. 
The highest-order isometry in this algebra is a screw motion (a simultaneous combination of translation and rotation) constructed using four reflections.
Crucially, we do not have to parameterize and compute four reflections explicitly.
Instead, we use the fact that four reflections lead to a scalar, six bivectors, and a quadvector component.
As such, we parameterize these components and refine them via gradient descent.
The points of the Tetris objects are implemented as \emph{trivectors}, representing intersections of three planes in a point.
That is, a point $x, y, z$ is implemented in $\sG_{3, 0, 1}$ as $xe_{023} + ye_{013} + ze_{012}$.

\textbf{GCA-MLPs. }
\Cref{fig:tetris_figure} summarizes various results of the Tetris experiment,
showing test mean squared errors (MSE) summed over the number of predicted time-steps as a function of the number of training trajectories.
The baseline MLP networks receive unstructured data in the form of a $3 \times 4 \times 4 \times 8 = 384$-dimensional vector, representing three coordinates per point, four points per object, and eight objects over four input time steps. 
The \emph{MotorMLP} also takes data in an unstructured manner but regresses directly on a rotation matrix and a translation offset, thereby also enforcing a geometric transformation.
However, this transformation is indirect, and the learned representations remain entirely unstructured.
We also consider two \emph{equivariant} baselines, both EMLP models as presented in \citet{finzi2021practical}.
Indeed, the task has global equivariance properties; the frame of reference does not affect the trajectory's unfolding.
However, \emph{local} motions are what makes the task challenging.
As such, we see that the $\O(3)$ and $\SO(3)$ equivariant models only marginally outperform the baseline MLPs.
In contrast, our GCA-MLP significantly outperforms the baseline MLPs.

\textbf{GCA-GNNs. }
Next, we test message-passing graph networks (GNN) \citep{gilmer2017neural, battaglia2018relational} where we encode the position coordinates as nodes and thus provide a strong spatial prior to the learned function.
We consider small (S) and large (L) versions, where the small ones have the same number of parameters as the baseline MLPs.
We incorporate versions that include the relative distances (+d) between the coordinates in the messages.
We further include the dynamic graph convolutional network \emph{EdgeConv} \citep{wang2019dynamic} in the baselines.
Finally, we introduce GCA-GNNs, which apply the same message-passing algorithm as the baseline GNNs, but replace message and node update networks with GCA-MLPs, i.e.,
\begin{align}
   \mv m_{ij} := \phi_e(\mv h^l_i, \mv h^l_j) \ , \ \ \ 
   \mv h^{l+1}_i := \phi_h(\bar{\mv m_i}, \mv h^l_i) \ , 
\end{align}
where $\mv h_i^l \in \sG_{3, 0, 1}^c$ are the node features at node $i$, $\mv m_{ij} \in \sG_{3, 0, 1}^c$ are messages between node $i$ and node $j$, and $\bar{\mv m}_i \in \sG_{3, 0, 1}^c$ is the message aggregation for each node. 
We put $\mv h_i^0:=\mv x_i$, where $\mv x_i$ are the point coordinates embedded in the algebra $\sG_{3, 0, 1}$.
The combination of graph structure and geometric transformations (GCA-GNNs) outperforms all the baselines, sometimes by a large margin.

Next, we show how we naturally combine objects in GCANs without sacrificing expressiveness.
We do so by including the discretized velocities of the points at all time steps as model inputs, predicting future coordinates and future velocities.
In GCANs, we can include the velocities directly as vector components $e_1$, $e_2$, and $e_3$ alongside the trivector components that encode the object positions, and use the exact same neural network architecture to transform them. 
Effectively, having more components increases the number of numerical operations used to compute \Cref{eq:ga_linear}. However, the number of parameters stays the same, and weights are now shared between multivector components.
Consequently, both attributes then are simultaneously subjected to Euclidean rigid motions throughout the network.
In contrast, the baseline models must account for the additional input and output of velocity data, requiring a reduction in the size of the hidden layers to maintain the same number of parameters.

\subsection{Fluid mechanics}
\begin{figure*}
    \centering
    \includegraphics[width=\linewidth]{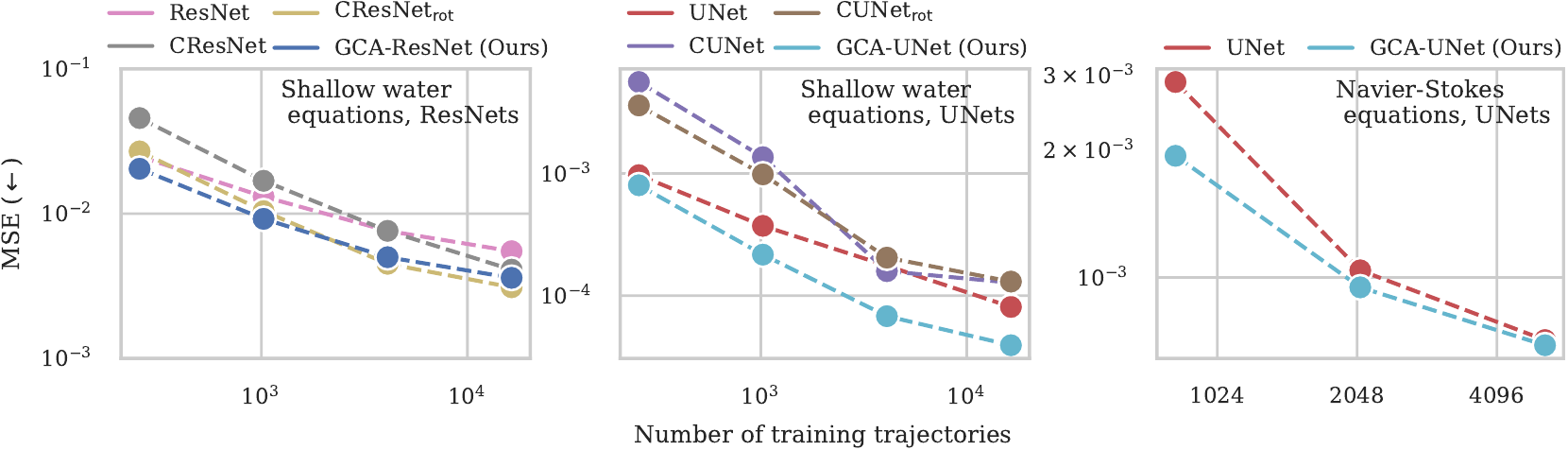}
    \caption{MSE results of the large-scale fluid mechanics experiments as a function of the number of training trajectories. We compare ResNet (left) and UNet (center) models on the shallow water equations. Right: UNet comparison on the Navier-Stokes equations.}
    \label{fig:results_fluid_dynamics}
\end{figure*}
In the following experiments, we aim to learn large-scale partial differential equation (PDE) surrogates on data obtained from numerical solvers.
To be more precise, we aim to learn the mapping from some fields, e.g., velocity or pressure fields, to later points in time. 
In this work, we investigate PDEs of fluid mechanics problems. 
Specifically, we focus on the $2+1$-dimensional shallow water~\citep{vreugdenhil1994numerical} and the $2+1$-dimensional incompressible Navier-Stokes equations~\citep{temam2001navier}.

\begin{figure}
\centering
\includegraphics[width=\linewidth]{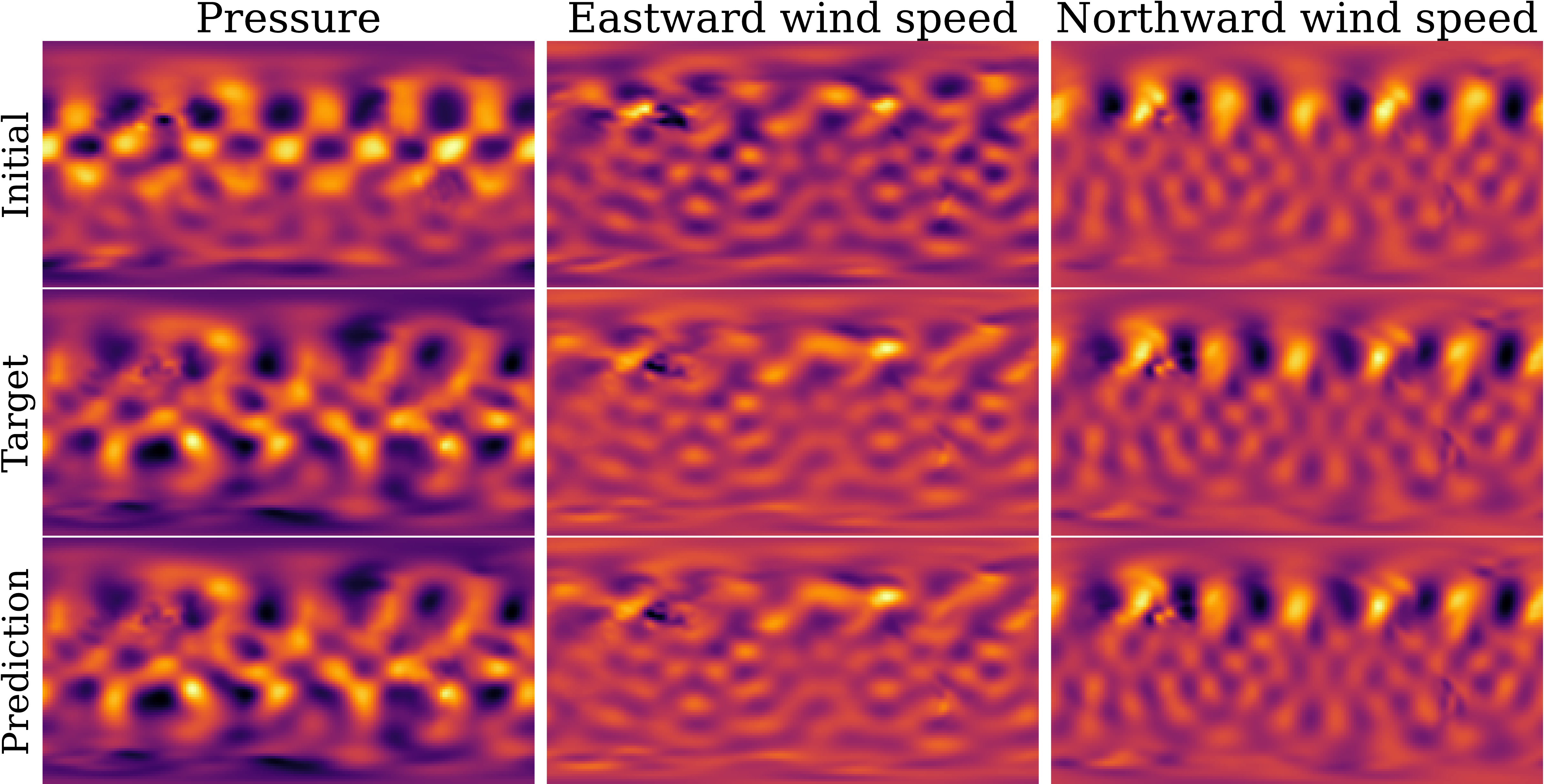}
\caption{Example input, target, and predicted fields for the shallow water equations. Predictions are obtained by the GCA-UNet model when using 16384 training trajectories.
}
\label{fig:shallow-water-fields}
\end{figure}

\textbf{Shallow water equations. }
The shallow water equations describe a thin layer of fluid of constant density in hydrostatic
balance, bounded from below by the bottom topography and from above by a free surface. 
As such, the shallow water equations consist of three coupled PDEs, modeling the temporal propagation of the fluid velocity in $x$- and $y$-direction, and the vertical displacement of the free surface, which is used to derive a scalar pressure field. 
For example, the deep water propagation of a tsunami can be described by the shallow water equations, and so can a simple weather model.  
We consider a modified implementation of the \texttt{SpeedyWeather.jl} \citep{milan_klower_2022_6788067} package, obtaining data on a $192\times96$ periodic grid ($\Delta x=1.875^\circ$, $\Delta y=3.75^\circ$) with temporal resolution of $\Delta t=6$ h. 
The task is to predict velocity and pressure patterns $6$ hours into the future given four input time steps.
Example input and target fields are shown in \Cref{fig:shallow-water-fields}.

\textbf{GCA-CNNs. }
When building GCA-CNNs, we use the fact that scalar pressure and vector velocity field are strongly coupled in the underlying shallow water equations.
We, thus, consider them as a single entity in a (higher-dimensional 3D) vector space and assume that it transforms under scaling and rotation.
We embed data in $\sG_{3,0,0}$ as vectors, constructing 2D convolutional GCA layers of the form 
\begin{align}
    \mv x &\mapsto \sum_{i=1}^{c \times k \times k} w_i \cdot \mv a_i \mv x_i \mv a_i^{-1} 
    \ , \label{eq:ga_conv}
\end{align}
where $k$ denotes a pre-specified kernel size, and $c$ is the number of input channels.
The group action weights $\mv a_i$ are equipped with nonzero scalar and bivector components, yielding rotations.
In \Cref{supp:speedyweather_appendix} we show that in this case, \Cref{eq:ga_conv} 
resembles the rotational Clifford layer introduced in~\citet{brandstetter2022clifford}.
We compare our methods against residual networks (ResNet)~\citep{he2016deep}, and modern UNet architectures \citep{ronneberger2015u, ho2020denoising},
which are considered to be the best-performing models for the task at hand~\citep{gupta2022towards}.
We replace their linear layers with layers of the form of \Cref{eq:ga_conv}.
Next, we directly replace normalization and nonlinearities with our proposed versions.
Altogether, this leads to GCA-ResNet and GCA-UNet architectures.
Further, we include the Clifford algebra versions of both models as presented in \citet{brandstetter2022clifford}. 
All models are optimized using similar numbers of parameters, i.e., 3 and 58 million parameters for ResNet and UNet architectures, respectively.
Hyperparameter choices and further details are summarized in \Cref{supp:speedyweather_appendix}.
The results are shown in \Cref{fig:results_fluid_dynamics}, reporting the mean-squared error (MSE) loss at a target time step summed over fields.

For ResNets (\Cref{fig:results_fluid_dynamics} left), we observe a similar overall picture as reported in~\citet{brandstetter2022clifford}.
Our GCA-ResNet and the conceptually similar rotational Clifford ResNet perform best where the overall performance is weak compared to UNet models. The reason is that ResNets as backbone architectures struggle to resolve the local and global processes of PDEs at scale.

For UNets (\Cref{fig:results_fluid_dynamics} center), we observe substantial performance gains of our GCA-UNets over baseline architectures, which for larger numbers of training samples is more than a factor of 5. We attribute
those performance gains to the strong inductive bias introduced via the geometric template idea.
More concretely, for larger backbone architectures such as UNets, which have different resolution, normalization, and residual schemes, it seems crucial that the map from $k$-vectors to $k$-vectors through layers and residual blocks is preserved. We thus show that GCA ideas scale to large architectures.
An exemplary qualitative result is shown in \Cref{fig:shallow-water-fields}.

\textbf{Navier-Stokes equations. }
Finally, we test the scalability of our models on a Navier-Stokes large-scale PDE experiment with a scalar (smoke density) field and a velocity field.
The scalar smoke field is \emph{advected} by the vector field, i.e., as the vector field changes, the scalar quantity is transported with it. 
The equations and simulations are implemented using the $\Phi$\texttt{Flow} \citep{holl2020learning} package.
The grid size for the simulation is $128 \times 128$, and the temporal resolution is $1.5$s.
Similar to the shallow water equations experiment, we embed the scalar and vector components as a single entity in the algebra $\sG_{3, 0, 0}$ that transforms under rotation and scaling.
We employ a convolutional layer similar to \Cref{eq:ga_conv}.
The results are shown in \Cref{fig:results_fluid_dynamics}, reporting the mean-squared error (MSE) loss at a target timestep summed over fields.
We include the UNet baseline and the respective GCA version.
In contrast to the shallow water equations, the coupling between the fields is less pronounced in this experiment. 
Therefore, our geometric interpretation has a weaker grounding.
However, as in the previous experiments, the geometric templates of GCA-UNets prove beneficial.

\section{Conclusion}
We proposed Geometric Clifford Algebra Networks (GCANs) for representing and manipulating geometric transformations.
Based on modern plane-based geometric algebra, 
GCANs introduce group action layers and the concept of refineable geometric templates. We showed that GCANs excel at modeling rigid body transformations, and that GCANs scale well when applying them to 
large-scale fluid dynamics simulations. 

\textbf{Limitations and future work. }
The main limitation of GCANs is that,  
as observed in~\citet{hoffmann2020algebranets} and~\citet{brandstetter2022clifford}, the compute density of (Clifford) algebra operations might lead to slower runtimes and higher memory requirements for gradient computations. 
Although higher compute density is, in principle, advantageous for hardware accelerators like GPUs, obtaining such benefits can require custom GPU kernels or better modern compiler-based kernel fusion techniques~\citep{tillet2019triton,AnselTorchDynamo2022}.
But even without writing custom GPU kernels, we have already managed to reduce training times for convolutional $\sG_{3,0,0}$ layers by an order of magnitude to a factor of roughly $1.2$ relative to comparable non-GCAN layers using available PyTorch operations.
Possible future work comprises extending geometric templates to \emph{geometric object templates}, i.e., transforming whole objects, such as molecules, according to group actions.  
Further, the covariant transformation of objects in geometric algebra might be exploited to build equivariant architectures \cite{ruhe2023clifford}.
Finally, we aim to leverage the fact that geometric algebra gives us better primitives to deal with object geometry (mesh) transformations and corresponding fluid dynamics around those objects, as seen in, e.g., airfoil computational fluid dynamics \citep{bonnet2022airfrans}.

\section*{Acknowledgements}
We would like to sincerely thank Leo Dorst for providing feedback on the final version of the paper.
His work on geometric algebra has been instrumental in shaping our understanding of geometric algebra.

We also thank Patrick Forr\'e for helpful conversations on the formalization of Clifford algebras, 
and Markus Holzleitner for proofreading the manuscript.

\bibliography{bib}
\bibliographystyle{icml2023}

\appendix
\onecolumn
\section{Glossary}
In \Cref{tab:glossary}, we provide an overview of notations that are commonly used throughout the paper.
\begin{table}[h]
    \centering
    \begin{tabular}{@{}p{0.2\linewidth}p{0.8\linewidth}@{}}
\toprule
Notation               & Meaning                                                                                                                                                                                                     \\ \midrule
$\mathcal{G}$          & Group                                                                                                                                                                                                       \\
$\sG_{p, q, r}$        & A geometric algebra with $p$ positive dimensions, $q$ negative dimensions and $r$ zero dimensions. $q$ and $r$ are left out of the notation when they equal $0$.                                            \\
$\Pin(p, q, r)$        & $\Pin$ group with $p$ positive dimensions, $q$ negative dimensions and $r$ zero dimensions.                                                                                                                 \\
$u$, $v$, $w$          & Abstract $\Pin(p, q, r)$ group elements.                                                                                                                                                                             \\
$u_1 u_2 \dots u_k$    & A $k$-reflection (composition of $k$ reflections) Abstract element of $\Pin(p, q, r)$.                                                                                                                              \\
$u v u^{-1}$           & Group conjugation of the form $G \times G \to G$. The group acts on itself.                                                                                                                                                                                          \\
$\mv x$                & A multivector of $\sG_{p, q, r}$.                                                                                                                                                                           \\
$[\mv x]_k$            & Selects the $k$-vector part of $\mv x$. For example, $[\mv x]_0$ selects the scalar part, $[\mv x]_1$ the vector part and $[\mv x]_2$ the bivector part.                                                     \\
$\mv x \mv y$          & Geometric product between $\mv x$ and $\mv y$.                                                                                                                                                              \\
$\lambda \mv x$        & Scalar product of scalar $\lambda$ with multivector $\mv x$.                                                                                                                                                \\
$\mv u, \mv v$         & $\Pin(p, q, r)$ elements expressed in $\sG_{3, 0, 1}$.                                                                                                                                                       \\
$\mv u_1, \mv u_2, \dots \mv u_k$ & A $k$-reflection (composition of $k$ reflections using geometric products) expressed in $\sG_{3, 0, 1}$. \\
$(-1)^{kl} \cdot \mv u \mv v \mv u^{-1}$ & $\Pin(p, q, r)$ group action expressed in elements of $\sG_{3, 0, 1}$. \\
$\mv u^{-1}$           & Multiplicative inverse (using the geometric product) of $\mv u$. That is, $\mv u \mv u^{-1}=1$.                                                                                                                 \\
$\mv a_i$              & $\Pin(p, q, r)$ group element used as group action in our geometric algebra neural layers.                                                                                                                  \\ 
$e_i$                  & A basis vector of $\sG_{p, q, r}$.                                                                                                                                                                          \\
$e_i e_j$              & A basis bivector of $\sG_{p, q, r}$.                                                                                                                                                                        \\
$e_i e_j e_k$          & A basis trivector of $\sG_{p, q, r}$.                                                                                                                                                                       \\
$e_0$                  & Fourth (special) basis vector of a geometric algebra modeling three-dimensional space. $e_0^2=\{-1, 0, 1 \}$ leads to hyperbolic, projective and Euclidean geometry, respectively. \\
$I$                    & The pseudoscalar of a geometric algebra.                                                                                                                                                                    \\
$x, y, z$              & Axes of a (Euclidean) coordinate system.                                                                                                                                                                    \\
$\Spin(n)$             & Special $\Pin(n)$ group, excluding improper isometries.                                                                                                                                                     \\
$\O(n)$,               & $n$-dimensional orthogonal group.\\
$\SO(n)$               & $n$-dimensional special orthogonal group.\\
$\Euc(n)$              & $n$-dimensional Euclidean group.\\
$\SEuc(n)$             & $n$-dimensional special Euclidean group.\\
$X$                    & Vector space.                                                                                                                                                                                               \\
$\alpha(\cdot, \cdot)$ & Group action of our group action linear layers.                                                                                                                                                                                               \\
$g_i$                  & Group element.                                                                                                                                                                                              \\
$x_i$                  & Vector space element.                                                                                                                                                                                       \\
$w_i$                  & Scalar neural network weight.                                                                                                                                                                               \\ \bottomrule
\end{tabular}
    \caption{Overview of notations commonly used in the paper.}
    \label{tab:glossary}
\end{table}
\newpage

\section{Geometric Algebra}
\label{supp:ga}
Technically, there are no differences between geometric and (real) Clifford algebra \footnote{In fact, Clifford himself chose ``geometric algebra''.}.
However, it is common practice to use Clifford algebra when interested in mathematical concerns (e.g., beyond using real numbers), and geometric algebra when interested in geometry.
A Clifford algebra is constructed by equipping a vector space with a quadratic form (see \Cref{sec:supp_clifford}).
The number of positive ($p$), negative ($q$), or zero $(r)$ eigenvalues (usually representing dimensions) of the \emph{metric} of the quadratic form determines the signature of the algebra.
Specifically, for an $n$-dimensional real vector space $\mathbb{R}^n$ ($n=p+q+r$) we can choose a basis with
\begin{equation}
\begin{aligned}
    e_i^2&=+1 & \quad 1 \leq i \leq p,  \\
    e_i^2&=-1 & \quad p < i \leq p+q,  \\
    e_i^2&=0 & \quad p+q < i \leq n,  \\
    e_ie_j&=-e_je_i & i \neq j \,. \label{eq:clifford_relations}
\end{aligned}
\end{equation}
These identies originate from the fact that the geometric product of two vectors yields a quadratic form and an anti-symmetric wedge product (\Cref{sec:supp_clifford}).
For two parallel vectors, the wedge product is zero, meaning we only get the scalar quadratic form. 
For two orthogonal vectors, the scalar part is zero and we only obtain the anti-symmetric part.
The metric is usually diagonal with elements in $\{+1, -1, 0\}$, and we can similarly use unit basis vectors to produce all the identities of \Cref{eq:clifford_relations}.

After picking a basis for the underlying vector space $\mathbb{R}^n$, multiplying its components yields higher-order basis elements, called \emph{basis blades}.
Through this construction, the \emph{algebraic} basis of the algebraic vector space has $2^n$ elements.
For example, when $n=3$, the space is spanned by $\{1,  e_1,  e_2,  e_3,  e_{12},  e_{13},  e_{23},  e_{123} \}$, where $ e_{ij}$ is shorthand for $ e_i  e_j$.
Note that this set is closed under multiplication with elements from itself using the relations \cref{eq:clifford_relations}.
The \emph{grade}\footnote{Technically, a Clifford algebra is not a graded algebra.} of a blade is the dimensionality of the subspace it represents.
For example, the grades of $\{1,  e_1,  e_{12},  e_{123} \}$ are 0 (scalar), 1 (vector), 2 (bivector) and 3 (trivector), respectively. 
The highest grade basis blade $I := e_{1}\ldots e_{n}$ is also known as the \emph{pseudoscalar}.
A vector is written as $x_1 e_1 + x_2 e_2 + x_3 e_3$, a bivector as $x_{12} e_{12} + x_{13} e_{13} + x_{23} e_{23}$, and so on.
Similarly to how vectors can be interpreted as oriented line segments, bivectors can be interpreted as oriented plane segments and trivectors as oriented cube segments.
We can construct $k$-vectors ($k \leq n$) by homogeneously combining basis blades of grade $k$.
A \emph{multivector} $\mv x \in \sG_{p, q, r}$ is a sum of $k$-vectors, such that $\mv x = [\mv x]_0 + [\mv x]_1 + \dots + [\mv x]_n$, where $[\mv x]_k$ denotes the $k$-vector part of $\mv x$.
Combining $k$-vectors leads to the most general element of the algebra: a \emph{multivector}.
In a three-dimensional algebra, this takes the form
\begin{equation}
    \begin{aligned}
        \mv x = &\underbrace{x_0 1}_{\text{Scalar}} + \underbrace{x_1 e_1 + x_2 e_2 + x_3 e_3}_{\text{Vector}}  + \underbrace{x_{12}e_{12} + x_{13}e_{13} + x_{23}e_{23}}_{\text{Bivector}} + \underbrace{x_{123}e_{123}}_{\text{Trivector}} \ .
    \end{aligned}
\end{equation}

We usually write $\sG_{p, q, r}$ (or $Cl_{p, q, r}$) to indicate what algebra we are using, where we suppress from  $\sG(\mathbb{R}_{p, q, r})$ the underlying generating vector space argument. 
Sometimes, the $q$ and $r$ components are left out when equal to zero.
The specific choice of algebra (determined by $p, q$ and $r$) allows for efficient modeling many types of geometry.

\textbf{Clifford multiplication: the geometric product. }
Multiplying two elements of the algebra yields the geometric product.
It is associative,
\begin{align}
    (\mv x \mv y) \mv z = \mv x (\mv y \mv z)\,,
\end{align}
left and right distributive, 
\begin{align}
    \mv x (\mv y + \mv z) &= \mv x \mv y + \mv x \mv z \\
    (\mv x + \mv y) \mv z &= \mv x \mv z + \mv y \mv z\,,
\end{align}
closed under multiplication,
\begin{align}
    \mv x \mv y \in \sG_{p, q, r}\,,
\end{align}
communicative with scalars,
\begin{align}
    \lambda \mv x = \mv x \lambda\,.
\end{align}

Using the associativity and distributivity laws of the algebra, we get for two multivectors with $1$-vector components only, i.e., commonly known ``vectors'', 
$\mv u=u_1 e_1 + u_2 e_2$ and $\mv v=v_1 e_1 + v_2 e_2$\,,
\begin{align}
     \mv u  \mv v &= (u_1 e_1 + u_2 e_2)(v_1 e_1 + v_2 e_2) \\
    &= \langle \rvu, \rvv \rangle + u_1v_2 e_{12} + u_2v_1 e_{21} \\
    &= \langle \rvu, \rvv \rangle + (u_1v_2 - u_2v_1) e_{12} \\
    &= \langle \rvu, \rvv \rangle + \rvu \wedge \rvv \,,
\end{align}
where $\langle \cdot, \cdot \rangle$ and $\cdot \wedge \cdot$ are the quadratic form and wedge product by construction (\Cref{sec:supp_clifford}).

Let $g_{ij}:= \langle e_i, e_j \rangle$.
In general, we can compute the geometric product using its associativity and distributivity laws for two multivectors like
\begin{equation}
\begin{aligned}
    \mv x \mv y &= \left(x_0 y_0 + g_{11} x_1 y_1 + g_{22} x_2 y_2 - g_{11}g_{22} x_{12} y_{12} \right)  1  \\
    &+ \left( x_0 y_1 + x_1 y_0 - g_{22}x_2 y_{12} + g_{22}x_{12}y_2 \right)  e_1  \\
    &+ \left( x_0 y_2 + g_{11} x_1 y_{12} + x_2 y_0 - g_{11}x_{12}y_1 \right)  e_2  \\
    &+ \left( x_0 y_{12} + x_1 y_2 - x_2y_1 +x_{12}y_0\right)  e_{12}\,,
\end{aligned}
\end{equation}
where now $\mv x = x_0 1 + x_1 e_1 + x_2 e_2 + x_{12} e_{12}$ and $\mv y = y_0 1 + y_1 e_1 + y_2 e_2 + y_{12} e_{12}$.
This is the primary operation of \citet{brandstetter2022clifford}.

\textbf{Duality. }
We can divide a geometric algebra into the vector subspaces that are spanned by each $k$-vector.
As such, we get
\begin{align}
\sG_{n} = \sG_{n}^0 \oplus \sG_{n}^1 \oplus \dots \oplus \sG_{n}^n\,.
\end{align}
The dimensionality (number of basis blades) of $\sG_{n}^k$ is given by $\binom{n}{k}$.
\begin{figure}
    \centering
    \includegraphics[width=0.5\linewidth]{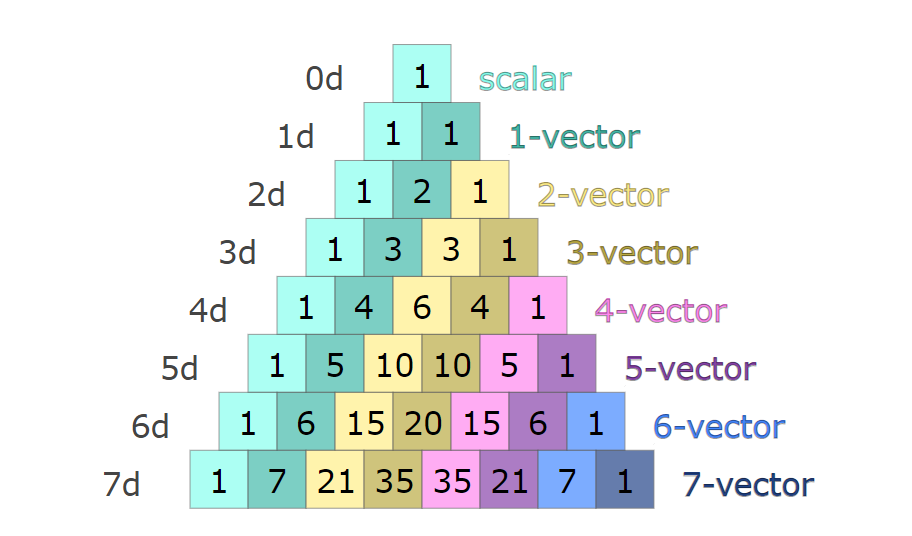}
    \caption{This figure shows the symmetric duality structure of geometric algebras. Up to seven dimensions, we show the number of $k$-blades up to the pseudoscalar ($n$-vector) for $n=0 \dots 7$. For each grade, we see that we have an equal number $\binom{n}{k} = \binom{n}{n-k}$ of dual blades.}
    \label{fig:duality}
\end{figure}
Note that $\binom{n}{k} = \binom{n - k}{k}$.
This symmetry shows the duality of the algebra (depicted in \Cref{fig:duality}).
Multiplying a multivector $\mv x $ with the pseudoscalar $I$ yields its dual $\mv x I$.
That is, scalars map to pseudoscalars, vectors to $(n-1)$-vectors, and so forth.

\textbf{Grade reversion and normalization. }
\label{supp:reversion_normalization}
Let $\mv u \in \sG_{p, q, r}$ be a $k$-reflection, i.e., $\mv u := \mv u_1 \dots \mv u_k$ where $\mv u_i$ are reflections (implemented as $1$-vectors).
We define the \emph{grade reversion} operator as 
\begin{align}
    \mv u^\dagger := \mv u_k \dots \mv u_1\,,
\end{align}
which is an involution $\sG_{p, q, r}\to \sG_{p, q, r}$.
In practice, this can be efficiently computed by noting that
\begin{align}
    [\mv u^\dagger]_k = (-1)^{k(k-1)/2} [\mv u]_k\,, \label{eq:grade_reversion}
\end{align}
where $[\mv u]_k$ selects the $k$-vector part of $\mv u$.
For example, $[\mv u]_2$ selects the bivector components of $\mv u$.
Effectively, this simply flips the sign of the components that have $k =2 \mod 4$ or $k=3 \mod 4$.
In geometric algebras of dimension $n \leq 3$, $\mv u^\dagger \mv u$ is a scalar quantity.
We can therefore define a norm 
\begin{align}
    \lVert \mv u \rVert := \sqrt{\lvert \mv u^\dagger \mv u \rvert }\,.
\end{align}
For normalized $\mv u$, we have
\begin{align}
    \mv u^\dagger \mv u = \mv u^{-1} \mv u = 1\,.
\end{align}

For example, using $\sG_{2, 0, 0}$ we have for $\mv u= \frac1{\sqrt 2} + \frac 1{\sqrt 2} e_{12}$, $\mv, \mv u^\dagger = \frac1{\sqrt{2}} - \frac1{\sqrt{2}}e_{12}$,  $g_{11}=1$, and $g_{22}=1$ we get 
\begin{align}
    \mv u\mv u^\dagger &= \left (\frac1{\sqrt 2} + \frac1{\sqrt 2}e_{12}\right)\left(\frac 1{\sqrt 2} - \frac1{\sqrt 2} e_{12}\right) \\
    &=\frac12 - \frac12 e_{12} + \frac12 e_{12} - \frac12 e_{12} e_{12} \\
    &= \frac12 + \frac 12 e_1^2 e_2^2   \\
    &= 1\,.
\end{align}

\textbf{Reflections. }
Using vector algebra, a reflection of $v \in \mathbb{R}^n$ in the hyperplane with normal $u$ is given by
\begin{align}
    v \mapsto v - 2 \frac{\langle u, v \rangle}{\langle u, u \rangle} u\,,
\end{align}
which, intuitively, subtracts from $v$ the projection of $v$ onto $u$ twice.
Using the geometric product and embedding the vectors as $1$-vectors, we can rewrite this as 
\begin{align}
    \mv v - 2 \frac{\langle \mv u, \mv v \rangle}{\langle \mv u, \mv u \rangle} \mv u &= \mv v - (\mv v \mv u + \mv u \mv v) \mv u^{-1} \\
    &= -\mv u \mv v \mv u^{-1}\,,
\end{align}
where we used $\mv u^{-1}:= \frac{\mv u}{\langle \mv u, \mv u \rangle}$ since 
\begin{align}
    \mv u \mv u^{-1} = \frac{\mv u^2}{\langle \mv u, \mv u \rangle } =1\,,
\end{align}
and we used the fundamental Clifford identity $\langle \mv u, \mv v \rangle = \frac12 (\mv u \mv v + \mv v \mv u )$ (for two vectors, \cref{sec:supp_clifford}).

The Cartan-Dieudonné theorem tells us that all higher-order orthogonal transformations of an $n$-dimensional space can be constructed from at most $n$ reflections.
As such, we can apply
\begin{align}
    \mv v \mapsto (-1)^{k} \mv u_1 \dots \mv u_k \, \mv v \, (\mv u_1 \dots \mv u_k)^{-1} = \mv u_1 \dots \mv u_k \, \mv v \, \mv u_{k}^{-1} \dots \mv u_1^{-1}
\end{align}
to compute a $k$-reflection.
We used the fact that  for normalized vectors $\mv u_i$, $(\mv u_1 \dots \mv u_k)^{-1} = (\mv u_1 \dots \mv u_k)^\dagger$, which can simply be computed using \Cref{eq:grade_reversion}.

\textbf{Outermorphism. }
A linear map $F: \mathbb{G}_n \to \mathbb{G}_n$ is called an \emph{outermorphism} if 
\begin{enumerate}
    \item $F(1) = 1$
    \item $F(\mathbb{G}_n^m) \subseteq \mathbb{G}_n^m$
    \item $F(\mv x \mv y) = F(\mv x) F(\mv y)$.
\end{enumerate}
Property 2 means that such a map is \emph{grade-preserving}.
Further,
\begin{theorem}
For every linear map $f: \mathbb{R}^n \to \mathbb{R}^n$ there exists a unique outermorphism $F: \mathbb{G}_n \to \mathbb{G}_n$ such that for $x \in \mathbb{R}^n, F(x)=f(x)$.
\label{thm:outermorphism}
\end{theorem}

It can be shown (e.g., \citet{macdonald2012linear, hestenes2012clifford}) that reflections implemented using geometric products directly extend to outermorphisms.
As such, by induction, we get that their compositions are outermorphisms, hence \Cref{thm:cartan} applies and we get grade-preserving isometries    operating independently on the geometric algebra subspaces.

\textbf{Even subalgebras. }
We can construct \emph{even subalgebras} by considering only the basis blades that have $k=0 \mod 2$.
In certain cases, we can use subalgebras to model higher-dimensional transformations.
For example, $\sG_{0, 1}$ is isomorphic to the complex numbers $\sC$, but so are the even grades of $\sG_{2}$. 
We let the scalar part of $\sG_{2}$ identify with the real part of a complex number, and the bivector (pseudoscalar) part with the imaginary part. 
The bivector also squares to $-1$: $e_{12}e_{12} = -e_1^2 e_2^2 = -1$.
Similarly, $\sG_{0, 2}$ is isomorphic to the quaternions $\mathbb{H}$, which are often used to model three-dimensional rotations, but so are the even grade elements of $\sG_{3}$, i.e., the scalar and the three bivectors.
This can easily be seen, since bivectors square to $-1$ in $\sG_{3}$,
and so do vectors in $\sG_{0,2}$.
We usually say that bivectors parameterize rotations.

\textbf{Advantages of geometric algebra over vector algebra. }
\label{sec:supp_ga_advantages}
We discuss some advantages of geometric algebra over classical (vector) linear algebra. 
First and foremost, geometric algebra is an extension (completion) of vector algebra.
For example, cross-products, norms, determinants, matrix multiplications, geometric primitives, projections, areas, and derivatives can all be computed in geometric algebra in a (sometimes more) interpretable way.
Second, in vector algebra, we need several different approaches to represent basic geometric objects.
When done naively, this can lead to ambiguities. 
For example, one uses vectors for both directions and points.
In geometric algebra, one can naturally represent these objects as invariant subspaces.
Third, in vector algebra, we also need to implement transformations differently depending on the object.
In contrast, in geometric algebra, we parameterize transformations of space, and the objects transform covariantly, regardless of what they are.
Fourth, matrices are dense and hard to interpret.
That is, it is not straightforward to quickly see whether a certain matrix is, e.g., a rotation or reflection.
In geometric algebra, by just observing what components are parameterized, we can directly see what transformation a certain multivector parameterizes.
Finally, geometric algebra generalizes across dimensions. 
That is, a computational geometry computer script that works for, e.g., two dimensions also works for three and higher dimensions.
In other words, we do not need to call different functions depending on the dimensionality of the problem.
This is not usually the case for classical approaches.

\section{References}
In this section, we first discuss related scientific work and then provide further references to other matters involving (incorporating machine learning in) geometric algebra.

\subsection{Related Work}
We discuss work related to incorporating geometric priors in neural networks for dynamical systems, using orthogonal transformations or isometries in neural networks, Clifford or geometric algebras in deep learning.

\textbf{Geometric priors in dynamical systems. }
The use of machine learning, especially deep learning, has proven to be highly effective in tackling complex scientific problems \citep{li2020fourier, lam2022graphcast, jumper2021highly, sanchez2020learning, mayr2021boundary, thomas2018tensor, miller2021truncated, lippert2022learning, boelrijk2021bayesian, pandeva2022multi, hoogeboom2022equivariant, brandstetter2022message, brandstetter2022lie, ruhe2022detecting, ruhe2022normalizing, smidt2021euclidean, smidt2021finding, batzner20223}.
Many of these settings involve the dynamical evolution of a system.
These systems play out in spaces equipped with geometries, making geometry an essential aspect of modeling.
Geometric deep learning focusing on equivariance  \citep{bronstein2017geometric, bronstein2021geometric, cohen2016group, weiler20183d, finzi2021practical, geiger2022e3nn} or learning on manifolds \cite{feng2019meshnet, chakraborty2020manifoldnet, milano2020primal, koval2017statistical} forms a rich subfield enabling in which models can be parameterized such that they are either invariant of equivariant to group actions applied to the input features.
Methods arising from this philosophy have successfully been applied to scientific settings \citep{batatia2022mace, miller2020relevance, unke2021se, satorras2021n}.
For example, the celebrated AlphaFold protein folding architecture \citep{jumper2021highly} uses an $\Euc(3)$-equivariant attention mechanism.
There has been less focus on incorporating geometric priors in a similar sense to the current work: biasing a model towards (Euclidean) rigid motions, often found in dynamical systems.
Examples of inductive biases in dynamical systems are given by \citet{ruhe2021self}, who bias a particle's trajectory to transformations given by underlying prior physics knowledge of the system, and 
\citet{kofinas2021roto}, who provide local coordinate frames per particle in Euclidean space to induce roto-translation invariance to the geometric graph of the dynamical system. 

Regarding geometric priors in dynamical systems, we discuss works related to point clouds and neural partial differential equation surrogates.
Current leading work is provided by \citet{liu2019meteornet}, who build on \citet{qi2017pointnet}.
The work constructs spatiotemporal neighborhoods for each point in a point cloud.
Local features are thereby grouped and aggregated to capture the dynamics of the whole sequence.
\citet{cao2020asap} similarly propose a spatiotemporal grouping called Spatio-Temporal Correlation.
Further, \citet{byravan2017se3, behl2019pointflownet} are seminal works for modeling point cloud sequences.
There are several critical differences between the current work an the above-mentioned.
First, our approach is general in the sense that we can apply actions from several groups (e.g., $\O(n)$, $\SO(n)$ for all $n$, $\Euc(3)$, $\SEuc(3)$), depending on the application.
Second, we can naturally operate on objects of different vector spaces: (standard) vectors, points, lines, planes, and so on. And, most notably, we can operate on \emph{grouped} versions of these, transforming them in a coupled manner.
Third, we provide network layers that preserves the vector space of these objects.
Fourth, geometric algebra provides a flexible framework that can be extended to different kinds of geometry, such as \emph{conformal} geometry.
Finally, some of these do not explicitly provide a geometric prior, or their application setting is quite different (e.g., estimating Euclidean motion from images).
Works like \citet{liu2019flownet3d} and its extension (that includes a geometric prior in the form of a ``geometric loss function'') operate on voxalized grids, whereas we operate in this work directly on the objects.
\citet{wei2022spatial} introduce a spatiotemporal transformer that can enhance the resolution of the sequences.

Relatively less work has been published on incorporating geometry in neural PDE surrogates.
\citet{berg2018unified} propose to include partial derivatives with respect to space variables in the backpropagation algorithm.
\citet{gao2021phygeonet} propose \texttt{PhyGeoNet}, a convolutional network architecture supported by coordinate transformations between a regular domain and the irregular physical domain.
\citet{li2022fourier} similarly uses a coordinate transformation from irregular domains to a regular latent domain, which is now fully learned.
All of the above methods differ from the current approach by not incorporating geometric templates, the main contribution of the current work.
\citet{brandstetter2022clifford} propose a weaker version of geometric templates, which was discussed in detail throughout the current work.

\textbf{Orthogonality in neural networks. }
Next, we discuss more generally where orthogonal transformations or isometries have been used in deep learning.
For time series, \citet{mhammedi2017efficient, wang2020orthogonal} propose orthogonal recurrent neural networks through Householder reflections and Cayley transforms, respectively. 
The main contribution of orthogonality is a solution to exploding or vanishing gradients, and not necessarily incorporating geometry.
\citet{wang2020orthogonal} propose orthogonal filters for convolutional neural networks, also to enable training stability.

Further, orthogonality enables easy invertibility and cheap Jacobian determinants for normalizing flows \citep{berg2018sylvester, hoogeboom2020convolution}.

\textbf{Clifford and geometric algebras in neural networks. }
Incorporating Clifford algebras in neural networks is a rich subfield that started already with \citet{pearson1994neural, pearson2003clifford}.
Most of such approaches incorporate ``Clifford neurons'' \citep{vallejo2008clifford, buchholz2008clifford} as an extension of real-valued neural networks. 
For an overview of many of these works, consider \citet{brandstetter2022clifford}. 
More recent applications of Clifford algebras in neural networks are \citet{kuroe2011models, zang2022multi, trindade2022clifford}.
Most works that incorporate geometric algebra in neural networks are actually (in this sense) Clifford neural networks \citep{bayro1997geometric, bayro2001geometric}.
In the current work, we are more rigorous in building geometry into the models.
In particular, \citet{bayro2001geometric} (to our knowledge) were the first to propose taking linear combinations of geometric products with learnable scalar weights.
This contrasts similarly to previously mentioned works in the sense that the multivector grades completely get mixed.
Finally, \citet{spellings2021geometric} also use geometric algebra in a more principled way but use it to construct equivariant neural networks that can operate on geometric primitives other than scalars.

\subsection{Resources}
Excellent introductory books to geometric algebra for computer science or physics include \citet{hestenes2012new, doran2003geometric, dorst2010geometric,macdonald2012linear}.

\citet{keninckga, keninckdual, dorsrtcgi} provide outstanding introductory lectures to plane-based geometry and (projective and conformal) geometric algebra.
Further, \citet{nozickgeometric} give a talk about geometric neurons, and \citet{lopesclifford} discuss  
Clifford CNNs for lymphoblast image classification.

To get started with Clifford or geometric algebra in Python, \citet{python_clifford} is the go-to package.
Further, \citet{python_tfga} provides an implementation of many of these concepts in \texttt{TensorFlow}. \citet{python_jaxga} is its followup, where the same procedures are implemented in \texttt{JAX}.

Other great references include \citet{bivector.net} and \citet{sudgylacmoe}.

\section{Implementation Details}
In this section we discuss how we practically implemented our models and some practicalities like weight initialization.
To be complete, we repeat the definition of the group action linear layer using geometric algebra:
\begin{align}
    T_{g, w}(\mv x)=\sum_{i=1}^c w_i \cdot  \mv a_i \mv x_i \mv a_i^{-1}, 
\end{align}
with now $\rvx_i \in X:=\sG_{p, q, r}$, and $\rva_i \in \gG:=\Pin(p, q, r)$.

\subsection{Initialization}
As shown in \Cref{eq:scalar_linear_layer}, group action layers generalize scalar neural layers.
Therefore, we can get similar variance-preserving properties by using initialization schemes for $w=(w_1, \dots, w_c)$ from \citet{he2015delving, glorot2010understanding}, especially when considering that the group actions considered in this work are isometries.
Indeed, upon initialization, $\mv a_i$ are normalized, such that they preserve distances.
However, by freely optimizing them, we can get transformations that do scale the input by a small amount (to be precise, the squared norm of $\mv a_i$).
In a sense, we thus get an overparameterization of the scaling since $w_i$ also accounts for that.
However, we empirically observe that having free scalars explicitly parameterizing the linear combination of \Cref{eq:ga_linear} can yield stabler learning.
Further experimentation is needed to determine exactly why and when this is the case.

\subsection{GCA-MLPs}
\textbf{GCA linear layers. }
To parameterize a GCA-MLP, we first specify in which algebra $\sG_{p, q, r}$ we work and which blade components of the action $\mv a_i \in \sG_{p, q, r}$ we want to parameterize.
Further, similar to regular MLPs, we specify the number of input and output channels per layer.
The geometric algebra linear layer \Cref{eq:ga_linear} then first embeds data in the algebra.
For example, a vector in $\mathbb{R}^n$ gets embedded as $1$-vector components in $\sG_{p, q, r}$, leaving the rest of the blades to zero.
The sandwich $\mv a_i \mv x_i \mv a_i^{-1}$ consists of two geometric products, which can efficiently be implemented as matrix multiplications.
After this, we take a linear combination with the scalar weights $w_i$, which is straightforward to implement.

\textbf{MSiLU. }
In this work we use linear combinations of basis blades ($f_k(\mv x)$) in our nonlinearity
\begin{align}
     [\mv x]_k \mapsto \mathrm{MSiLU}_k\left(\mv x\right) := \sigma \left(f_k(\mv x)\right) \cdot   [\mv x]_k\,,
\end{align}
where $f_k(\mv x): \sG_{p, q, r} \to \mathbb{R}$ and $\sigma$ is the logistic function.
This can be efficiently implemented as a linear layer in all modern deep learning frameworks.

\textbf{Normalization. }
Finally, our normalization layer
\begin{align}
    [\rvx]_k \mapsto \frac{[\rvx]_k - \E[[\rvx]_k]}{\E[\lVert [\rvx]_k \rVert]} \ .
\end{align}
can be computed in a straightforward manner by computing an empirical average, and dividing by the average norm which is computed as described in \Cref{supp:reversion_normalization}.

\textbf{Architectures. }
For both the baseline MLP and GCA-MLP we use two hidden layers.

\subsection{GCA-GNNs}
We use the message propagation rules of \citet{gilmer2017neural, battaglia2018relational}.
Using multivectors, this amounts to
\begin{align}
    \mv m_{ij} &= \phi_e (\mv h_i^l, \mv h_j^l) \nonumber \\
    \bar{\mv m}_i &= \sum_{j \in \mathcal{N}(i)} \mv m_{ij} \\
    \mv h_i^{l+1} &= \phi_h(\mv h_i^l, \mv m_i) \nonumber \,,
\end{align}
where $\mv h_i^l \in \sG_{p, q, r}^c$ are the node features at node $i$, $\mv m_{ij} \in \sG_{p, q, r}^c$ are messages between node $i$ and node $j$, and $\bar{\mv m}_i \in \sG_{p, q, r}^c$ is the message aggregation for each node.
$\phi_c$ and $\phi_h$ are GCA-MLPs. 
Therefore, the whole GCA-GNN is a geometric template.
For the Tetris experiment, we use a fully connected graph since all the nodes are correlated.
Further, we use \texttt{PyTorch Geometric} \citep{fey2019fast} to implement the message passing algorithm.

\textbf{Architectures. }
The baseline GNNs use the message passing updates as proposed by \citet{gilmer2017neural, battaglia2018relational}.
We use four message passing layers, where the message and update networks $\phi_e, \phi_h$ (the non-GCA version) are implemented with scalar linear layers and \texttt{LeakyReLU} \citep{xu2015empirical} activation functions. 
The graphs are fully connected over 32 nodes (8 objects, each consisting of four point coordinates).
That allows for 12 input features (four input time-steps with 3 position values) for each node.
We further have embedding MLPs and output MLPs that map from the 12 input features to a number of hidden features and back.

The GCA-GNNs replace all the baseline modules with there GCA counterparts.
By coupling the $x, y, z$ coordinates in single multivector entities, we now only have $4$ input features.
Again, to account for the additional parameters that the group actions introduce, we reduce the number of hidden channels.

\subsection{GCA-ResNets}
Let $y \in \mathbb{Y} \subset \mathbb{Z}^2$ be a two-dimensional spatial coordinate in the domain $\mathbb{Y}$. 
A convolution operation\footnote{In deep learning, a convolution operation in the forward pass is implemented as cross-correlation.} convolves $x(y): \mathbb{Y} \to \mathbb{R}^{k \times k \times c}$, with a filter map where we extract a $k \times k$ ``patch'' around the coordinate $y$ with $c$ channels:
\begin{align}
     x \mapsto \sum_{i=1}^{c \times k \times k} w_i \cdot x_i \ ,
\end{align}
where we suppress the $y$ arguments.
We repeat this procedure using different weights depending on the specific output channel.

Now, we discuss the geometric algebra analog of the above. 
Let $\mv x(y): \mathbb{Y} \to \sG_{p, q, r}^{k \times k \times c}$ be a multivector-valued feature map where we extract a $k \times k$ patch around the coordinate $y$ with $c$ channels.
A group action convolution is then computed by
\begin{align}
    \mv x \mapsto \sum_{i=1}^{c \times k \times k} w_i \cdot \mv a_i \mv x_i \mv a_i^{-1} \ ,
    \label{eq:app_GCAN-CNN}
\end{align}
where we suppressed the spatial argument $y$.
It is important to note that in \Cref{eq:app_GCAN-CNN}, the weights $w_i \in \R$, but the weights $\rva_i \in \sG_{p, q, r}$. That is the geometric transformations given by the sandwich product $\rva_i \rvx_i \rva_i^{-1}$ are linearly mixed by $w_i$ afterwards.

\textbf{Architectures. }
We use the same ResNet implementations as \citet{brandstetter2022clifford}.
That is, we have 8 residual blocks, each consisting of two convolutional layers with $3 \times 3$ kernels (unit strides), shortcut connections, group normalization, and GELU activation functions \citep{hendrycks2016gaussian}.
Further, there are two embedding and two output layers.
The input time steps and fields are all encoded as input channels.
For example, for $4$ input time steps and $3$ input feature fields, we get 12 input channels.
Since the output space is the same as the input space, we do not need downsampling layers.
The CGA-ResNets directly replace each convolutional layer with a CGA layer, each GELU activation with MSiLU, and normalization with our normalization layer.
To keep the weights similar, we have to reduce the number of feature channels (see \Cref{sec:tetris_hyperparameters}).

\subsection{GCA-UNets}
\citet{gupta2022towards} show excellent performance of modern UNets on large-scale PDE surrogate tasks.
We use similar architectures.
Specifically, we encode the time steps and input feature fields as input channels to the model.
We use convolutional embedding and projection layers to upsample (downsample) the input (output) channels.
Throughout the network, we use $3 \times 3$ convolutions unit strides. 
We then use a channel multiplier structure of $(1, 2, 2, 2)$ which, similar to, e.g., \citet{saharia2022photorealistic}, shifts the parameters from the high-resolution blocks to the low-resolution blocks, increasing model capacity without encountering egregious memory and computation costs. 
At each resolution, we have a residual block and up- or downsampling layer. 
A residual block consists of two convolutions, activations (GELU), normalizations (Group Normalization), and a skip connection.
The bottleneck layer has two such residual layers.

For GCA-UNets, Clifford UNets, and rotational Clifford UNets, we, identically to the ResNet case, directly replace the layers, activations, and normalizations by the ones proposed in this paper and in \citet{brandstetter2022clifford}.

\section{Experiments}
\subsection{Computational Resources}
For the Tetris experiments, we used $1\times 40$ GB NVIDIA A100 machines.
The average training time for these experiments was 4 hours. 
The GCA-GNN took roughly 24 hours. 
However, as indicated before, several quick wins can be found to significantly reduce this.
For the fluid dynamics experiments, we used $2 \times 4 \times 16$ GB NVIDIA V100 machines.
The average training time was 24 hours.
The GCANs, in this case, did not perform significantly worse than the baselines. 
On the other hand, the Clifford neural networks, due to their expensive normalization procedure, took significantly longer to optimize.

\subsection{Tetris Experiment}
\textbf{Data generation. }
We take the shapes as provided by \citet{thomas2018tensor} and center them at the origin.
For every Tetris shape, we randomly sample a  rotation axis with a maximum angle of $0.05 \cdot 2 \pi$ and translation directions with a maximum offset of $0.5$.
This leads to a rotation matrix $\b P$ and translation vector $\b t $.
The position at time step $t$ ($t=0 \dots 8$) can then be computed by applying $\b P^t$ and $t \cdot \b t$ to an initial coordinate.
The discretized velocities are calculated by taking differences between the consecutive time steps.
We further add slight deformations to the structures by adding small Gaussian noise.

\textbf{Objective and loss function. }
After constructing a dataset of $N_{\text{train}}$ such trajectories, we predict, given four input time steps, the next four input timesteps and compare them against the ground-truth trajectories for all objects.
We define a loss function 
\begin{align}
    \mathcal{L}_{\text{MSE}} := \frac{1}{N_{\text{locations}}} \sum_{t=1}^{N_t} \sum_{y=1}^{N_{\text{locations}}}\sum_{p=1}^{{N_p}} \left( x_{typ} - \hat{x}_{typ} \right)^2
\end{align}
that expresses for all train datapoints and for all time steps the discrepancy between the predicted three-dimensional locations and the ground-truth ones.
When we predict positions, $N_p=3$.
When we also predict the velocities, we include those as well (i.e., then $p$ sums to 6).
In our experiments we have $N_t=4$ predicted output time steps and $N_\text{locations}=32$.
We average this loss over $1024$ validation and test trajectories.

\textbf{GCAN implementation. }
In this experiment, we consider the geometric algebra $\sG_{3, 0, 1}$.
The highest-order proper isometry is a screw motion (a simultaneous rotation and translation), which is implemented using the components $\mv a := a_0 1 + a_{01} e_{01} + a_{02} e_{02} + a_{03} e_{03} + a_{12} e_{12} + a_{13} e_{13}  + a_{23} e_{23} + a_{0123} e_{0123}$, where the coefficients are free parameters to be optimized.
Since points in geometric algebra are defined by intersection of three planes.
As such, we encode the $x, y, z$-coordinates of a point as $\mv x:= x e_{012} + y e_{013} + z e_{023}$ and transform them using group actions $\mv a$.
The $e_{123}$ (the dual of $e_0$) parameterizes a distance from the origin, as such we get $\mv x:= x e_{012} + y e_{013} + z e_{023} + \delta e_{123}$, where we leave $\delta$ in each layer as a free parameter.
The full input representation to the GCA-MLP has $N_{\text{locations}} \cdot N_t$ channels.
In contrast: a naive MLP has $N_{\text{locations}} \cdot N_t \cdot N_p$ input channels.
The GCA-MLP then transforms this input using geometric algebra linear layers.

The GCA-GNN additionally encodes the positions as graph nodes and therefore only has $N_t$ input channels.
Each message and each node update network of the message passing layers is implemented as a GCA-MLP as described above.

When we include the velocities then $\mv x := v_x e_1 + v_y e_2 + v_z e_3 + x e_{0123} + y e_{013} + z e_{023}$.
Importantly, we can use the same weights $\mv a$ to transform this multivector. However, to incorporate the knowledge of the additional velocities into the neural network weights, retraining is required.

\textbf{Model selection and optimization. }
\label{sec:tetris_hyperparameters}
We trained for $2^{17}=131072$ steps of gradient descent for all data regimes as reported in the main paper.
We did not run hyperparameter searches for these models and set them up using reasonable default architectures and settings.
Specifically, for the MLP baseline we use 2 layers of 384 hidden features resulting in 444K parameters.
We implement the $\O(3)$ and $\SO(3)$ baseline following~\citet{finzi2021practical}, where input and output representation are $8 \times 4 \times 4 T_1$ ($T_1$ is the vector representation).
The $\O(3)$, $\SO(3)$ and MotorMLP baselines have 256 channels each to obtain equal parameters.
The GCA-MLP has 128 channels to equal the parameter count.
The GNN baselines use four layers of message passing with 136 (small models, resulting in $444$K parameters) or 192 (large models, resulting in $893$K parameters) hidden features.
The EdgeConv baseline uses $256$ hidden features, resulting in 795K parameters.
The GCA-GNN equals the parameters of the small GNNs, using four hidden layers of 64 hidden features.

For the velocity experiment, the MLP baseline uses 2 hidden layers with 248 features to obtain the same parameter count as the large GNN baseline, i.e., $893$K.
The GNN baseline has four message passing layers with 192 hidden features, resulting in 900K parameters.
The GCA-MLP and GCA-GNN use the same number of hidden features as in the positions experiment, i.e., $444$K.

For all models we use the Adam \citep{kingma2014adam} optimizer with the default parameter settings (i.e., a learning rate of $10^{-3}$). We did no further extensive hyperparameter or architecture searches for these models and kept them to reasonable default settings.

\subsection{Extended results} 
In \Cref{tab:mse_tetris_mlp} we present numerically the results on the Tetris experiment for MLP-style models (also presented in \Cref{fig:tetris_figure}). 
\begin{table}[H]
\centering
\begin{tabular}{@{}lllll@{}}
\toprule
Training trajectories & 256             & 1024            & 4096                         & 16384                        \\ \midrule
MLP          & 4.5061          & 1.2180          & 0.2184                       & 0.1596                       \\
Motor-MLP    & 3.5732          & 0.8655          & 0.4249                       & 0.4020                       \\
$\SO(3)$-MLP & 1.5153          & 0.9504          & 0.2172                       & 0.1176                       \\
$\O(3)$-MLP  & 1.4232          & 0.7224          & 0.1584                       & 0.0876                       \\
GCA-MLP (Ours)      & \textbf{0.9852} & \textbf{0.0420} & \textbf{0.00732} &  \textbf{0.0061} \\ \bottomrule
\end{tabular}
\caption{Mean squared error of MLP-style models on the tetris experiment.}
\label{tab:mse_tetris_mlp}
\end{table}
In \Cref{tab:mse_tetris_gnn} we present numerically the results on the Tetris experiment for GNN-style models (also presented in \Cref{fig:tetris_figure}). 
\begin{table}[H]
\centering
\begin{tabular}{@{}lllll@{}}
\toprule
Training trajectories & 256    & 1024   & 4096                & 16384               \\ \midrule
EdgeConv     & 1.572  & 0.2406 & 0.0240              & 0.0123              \\
GNN (S)      & 0.3012 & 0.0432 & 0.0032              & 0.0020              \\
GNN (S, +d)  & 0.2887 & 0.0408 & 0.0040              & 0.0019              \\
GNN (L)      & 0.2879 & 0.0504 & 0.0043              & 0.0017              \\
GNN (L, +d)  & 0.2793 & 0.0516 & 0.0030              & 0.0016              \\
GCA-GNN (Ours)      & \textbf{0.2403} & \textbf{0.0012} & $\mathbf{6.1} \cdot \mathbf{10}^{-4}$ & $\mathbf{5.4} \cdot \mathbf{10}^{-4}$ \\ \bottomrule
\end{tabular}
\caption{Mean squared error of GNN-style models on the tetris experiment.}
\label{tab:mse_tetris_gnn}
\end{table}
In \Cref{tab:mse_tetris_gnn_velocity} we present numerically the results on the Tetris experiment where we include input and output velocities (also presented in \Cref{fig:tetris_figure}). 
\begin{table}[H]
\centering
\begin{tabular}{@{}lllll@{}}
\toprule
Training trajectories & 256    & 1024   & 4096                & 16384               \\ \midrule
MLP          & 3.2403 & 0.5040 & 0.1560              & 0.0912              \\
GCA-MLP      & 1.6560 & 0.0984 & 0.0418              & 0.0432              \\
GNN          & 0.0504 & 0.0086 & 0.0015              & $8.8\cdot 10^{-4}$  \\
GCA-GNN (Ours)      & \textbf{0.0122} & \textbf{0.0022} & $\mathbf{4.1} \cdot \mathbf{10}^{-4}$ & $\mathbf{2.6} \cdot \mathbf{10}^{-4}$ \\ \bottomrule
\end{tabular}
\caption{Mean squared error of GNN-style models on the tetris experiment when including velocity inputs and outputs.}
\label{tab:mse_tetris_gnn_velocity}
\end{table}

\subsection{Speedyweather}
\label{supp:speedyweather_appendix}

\textbf{Shallow water equations. }
The shallow water equations describe a thin layer of fluid of constant density in hydrostatic
balance, bounded from below by the bottom topography and from above by a free surface. For example, the deep water propagation of a tsunami can be described by the shallow water equations, and so can a simple weather model.
The shallow water equations read:
\begin{align}
    \frac{\partial v_x}{\partial t} + v_x \frac{\partial v_x}{\partial x} + v_y \frac{\partial v_x}{\partial y} + g\frac{\partial \eta}{\partial x} & = 0 \ , \nonumber \\
    \frac{\partial v_y}{\partial t} + v_x \frac{\partial v_y}{\partial x} + v_y \frac{\partial v_y}{\partial y} + g\frac{\partial \eta}{\partial y} & = 0 \ , \nonumber \\ 
    \frac{\partial \eta}{\partial t} + \frac{\partial}{\partial x} \bigg[ (\eta + h) v_x \bigg] + \frac{\partial}{\partial y} \bigg[ (\eta + h) v_y \bigg] & = 0 \ ,
    \label{eq:shallow-water}
\end{align}
where $v_x$ is the velocity in the $x$-direction, or zonal velocity,
$v_y$ is the velocity in the $y$-direction, or meridional velocity,
$g$ is the acceleration due to gravity,
$\eta(x,y)$ is the vertical displacement of free surface, which
subsequently is used to derive pressure fields; $h(x,y)$ is the topography of the earth's surface. 

\textbf{Simulation. }
We consider a modified implementation of the \texttt{SpeedyWeather.jl}\footnote{\url{https://github.com/milankl/SpeedyWeather.jl}}\citep{milan_klower_2022_6788067} package, obtaining data on a $192\times96$ periodic grid ($\Delta x=1.875^\circ$, $\Delta y=3.75^\circ$).
This package uses the shallow water equations \citep{vreugdenhil1994numerical}, a specific instance of a Navier-Stokes fluid dynamics system, to model global weather patterns.
The temporal resolution is $\Delta t=6$ h, meaning we predict these patterns six hours into the future.
SpeedyWeather internally uses a leapfrog time scheme with a Robert and William's filter to dampen the computational modes and achieve 3rd oder accuracy. \texttt{SpeedyWeather.jl} is based on the atmospheric general circulation model \texttt{SPEEDY} in Fortran~\citep{molteni2003atmospheric,Kucharski2013}.

\textbf{Objective. }
Given four input time steps, we predict the next time step and optimize the loss function
\begin{align}
    \mathcal{L}_{\text{MSE}} := \frac{1}{N_y} \sum_{t=1}^{N_t} \sum_{n=1}^{N_{\text{fields}}} \sum_{y \in \mathbb{Y}} (x_{tny} - \hat{x}_{tny})^2,
\end{align}
where $\mathbb{Y}\subset \mathbb{Z}^2$ is the (discretized) spatial domain of the PDE, $N_t$ is the number of prediction time steps, and $N_{\text{fields}}$ is the number of field components.
For the shallow water equations we have two velocity components and one scalar component.
Here, $\hat{x}_{n, t, y}$ is the predicted value at output time step $t$ for field component $n$ and at spatial coordinate $y$.
$x_{n, t, y}$ is its ground-truth counterpart.
We consider only one-step ahead loss, where $N_t=1$, as empirically it has been found that this naturally generalizes to better rollout trajectories \citep{brandstetter2022clifford, gupta2022towards}.

\textbf{GCAN implementation. }
It can be shown that the $\sG_{3, 0, 0}$ geometric algebra linear layer $T_{g, w}(\mv x)=\sum_{i=1}^c w_i \cdot  \mv a_i \mv x_i \mv a_i^{-1}$ reduces to $T_{g, w}(\mv x)=\sum_{i=1}^c w_i \cdot  \b R_i \mv x_i$.
\begin{equation}
\begin{aligned}
    \mv a \mv x &= (a_0 + a_{12} e_{12} + a_{13}e_{13}+a_{23}e_{23})(x_1 e_1 + x_2 e_2 + x_3 e_3) \\
    &= a_0 x_1 e_1 + a_0 x_2 e_2 + a_0 x_3 e_3  \\
    &- a_{12} x_1 e_2 + a_{12} x_2 e_1 + a_{12} x_3 e_{123} \\
    &- a_{13} x_1 e_3 - a_{13} x_2 e_{123} + a_{13} x_3 e_1 \\
    &+ a_{23} x_1 e_{123} - a_{23} x_2 e_3 + a_{23} x_3 e_2,
\end{aligned}
\end{equation}
which reduces to
\begin{equation}
\begin{aligned}
    \mv a \mv x &= (a_0 x_1 + a_{12} x_{2} + a_{13}x_3) e_1 \\
    &+ (a_0 x_2 - a_{12}x_1 + a_{23}x_3) e_2 \\
    &+ (a_0 x_3 - a_{13}x_1 - a_{23}x_2) e_3 \\
    &+ (a_{12} x_3 - a_{13}x_2 + a_{23}x_1) e_{123}
\end{aligned}
\end{equation}
Using grade reversion (\Cref{supp:ga}), we get $\mv a^{-1} = a_0 - a_{12} e_{12} - a_{13}e_{13} - a_{23}e_{23}$.
Then,
\begin{equation}
\begin{aligned}
    \mv a \mv x \mv a^{-1} &= (a_0^2 x_1 + a_0 a_{12} x_2 + a_0 a_{13} x_3) e_1 \\
    &+ (a_0^2 x_2 - a_0 a_{12} x_1 + a_0 a_{23} x_3) e_2 \\
    &+ (a_0^2 x_3 - a_0 a_{13} x_1 - a_0 a_{23} x_2)e_3\\
    &+ (a_0 a_{12} x_3 - a_0 a_{13} x_2 + a_0 a_{23} x_1) e_{123} \\
    &+ (-a_{12}a_0 x_1 - a_{12}^2 x_2 - a_{12} a_{13} x_3) e_2 \\
    &+ (a_{12} a_0 x_2 - a_{12}^2 x_1 + a_{12} a_{23} x_3) e_1 \\
    &+ (-a_{12} a_0 x_3 + a_{12} a_{13} x_1  +a_{12} a_{23} x_2) e_{123} \\
    &+ (a_{12}^2 x_3 - a_{12} a_{13} x_2 + a_{12} a_{23} x_1) e_3 \\
    &+ (-a_{13}a_0 x_1 -a_{13} a_{12} x_2 - a_{13}^2 x_3) e_3 \\
    &+ (a_{13} a_0 x_2 - a_{13} a_{12} x_1 + a_{13} a_{23} x_3) e_{123} \\
    &+ (a_{13} a_0 x_3 - a_{13}^2 x_1 - a_{13} a_{23} x_2) e_1 \\
    &+ (-a_{13} a_{12} x_3 + a_{13}^2 x_2 - a_{13} a_{23} x_1) e_2 \\
    &+ (-a_{23} a_0 x_1 - a_{23} a_{12} x_2 - a_{23} a_{13} x_3) e_{123} \\
    &+ (-a_{23}a_0 x_2 + a_{23} a_{12} x_1 - a_{23}^2 x_3) e_3 \\
    &+ (a_{23} a_0 x_3 - a_{23} a_{13} x_1 - a_{23}^2 x_2) e_2 \\
    &+ (a_{23} a_{12} x_3 - a_{23} a_{13} x_2 + a_{23}^2 x_1) e_1,
\end{aligned}
\end{equation}
which again reduces to
\begin{equation}
\begin{aligned}
    \mv a \mv x \mv a^{-1} &= 
    (a_0^2 x_1 + 2 a_0 a_{12} x_2 + 2 a_0 a_{13} x_3  - a_{12}^2 x_1 + 2 a_{12} a_{23} x_3 - a_{13}^2 x_1 - 2 a_{23} a_{13} x_2 + a_{23}^2 x_1) e_1 \\
    &+ (a_0^2 x_2 - 2 a_0 a_{12} x_1 + 2 a_0 a_{23} x_3 - a_{12}^2 x_2 - 2 a_{12} a_{13} x_3 + a_{13}^2 x_2 - 2 a_{13} a_{23} x_1 - a_{23}^2 x_2) e_2 \\
    &+ (a_0^2 x_3 - 2 a_0 a_{13} x_1 - 2 a_0 a_{23} x_2  + a_{12}^2 x_3 -  2a_{12} a_{13} x_2 + 2 a_{12} a_{23} x_1 - a_{13}^2 - a_{23}^2 x_3) e_3
\end{aligned}
\end{equation}
where we see that the trivector components cancel!
Collecting the terms, we can define 
\begin{align}
\b R := \begin{bmatrix}
a_0^2 - a_{12}^2 - a_{13}^2 + a_{23}^2 & 2 a_0 a_{12} -2a_{23}a_{13}  & 2a_0a_{13} + 2a_{12}a_{23} \\
-2a_0a_{12} -2 a_{13}a_{23}& a_0^2 - a_{12}^2 + a_{13}^2 - a_{23}^2 & 2a_0a_{23} -2a_{12}a_{13} \\
 -2a_0a_{13} +2a_{12}a_{23} & -2a_0 a_{23} -2 a_{12}a_{13} &  a_{12}^2 + a_0^2 - a_{13}^2 - a_{23}^2
\end{bmatrix}
\end{align}

Here, $\b R_i$ is a rotation matrix that acts on the vector components of $\mv x_i$.
This is a much more efficient implementation than the sandwich operation, and resembles the rotational layer of \citet{brandstetter2022clifford} but does not implement a scalar part (which would go against the concept of a group action layer).

\textbf{Model selection and optimization. }
For ResNet architectures we consider 128 and 144 feature channels, the latter matching the number of parameters of the GCA-ResNets.
Out of these two, the best performing model was reported in the paper.
These channels are kept constant (apart from embedding and decoding layers) throughout the network.
The Clifford ResNets and GCA ResNet consider 64 channels.
These architectures then have roughly 3M parameters.
The UNet models consider 64 and 70 base channels, the latter matching the parameters of the GCA-UNet (58M).
The Clifford and GCA counterparts have 32 channels.
At 448 training trajectories, we tested all models accross two learning rates ($2 \cdot 10^{-4}$ and $5 \cdot 10^{-4}$), and different normalization schemes. Normalization turned out to be beneficial at all times.
We further closely followed the hyperparameter settings as reported in~\citet{gupta2022towards}\footnote{\url{https://microsoft.github.io/pdearena/}}.
Further, we tested for the GCA models whether a learned linear combination, a simple summation, or averaging in the MSiLU layers works best.
Summation and a learned linear combination turned out to be the most promising.

We used the Adam optimizer with the best performing learning rate, and use cosine annealing \citep{loshchilov2016sgdr} with linear warmup.
We trained at all data-regimes for 50 epochs.

\textbf{Extended results. }

In \Cref{tab:mse_sw_resnet} we present the mean squared error values for ResNet style models that are also presented in \Cref{fig:results_fluid_dynamics}.
\begin{table}[H]
\centering
\begin{tabular}{@{}lllll@{}}
\toprule
Training trajectories                & 112    & 224    & 448   & 892  \\ \midrule
ResNet               & 0.0248 & 0.01310 & 0.0076 & 0.0055 \\
CResNet              & 0.0457 & 0.01685 & 0.0076 & 0.0041 \\
CResNet$_\text{rot}$ & 0.0269 & 0.01040 & \textbf{0.0045} & \textbf{0.0031} \\
GCA-ResNet (Ours)    &\textbf{ 0.0204} & \textbf{0.00920 }& 0.0050 & 0.0036 \\ \bottomrule
\end{tabular}
\caption{Mean squared error of ResNet-style models on the shallow water equations experiment.}
\label{tab:mse_sw_resnet}
\end{table}

In \Cref{tab:mse_sw_unet} we present the mean squared error values for UNet-style models that are also presented in \Cref{fig:results_fluid_dynamics}.
\begin{table}[H]
\centering
\begin{tabular}{@{}lllll@{}}
\toprule
Training trajectories       & 112      & 224     & 448     & 892    \\ \midrule
CUNet              & 0.0056 & 0.0013 & $1.59 \cdot 10^{-4}$ & $1.28 \cdot 10^{-4}$ \\
CUNet$_\text{rot}$ & 0.0036 & $9.83 \cdot 10^{-4}$ & $2.05 \cdot 10^{-4}$ & $1.31 \cdot 10^{-4}$ \\
UNet               & $9.69\cdot 10^{-4}$ & $3.73 \cdot 10^{-4}$ & $1.76\cdot 10^{-4}$  & $8.10 \cdot 10^{-5}$ \\
GCA-UNet (Ours)    & $\mathbf{8.01}\cdot\mathbf{10}^{-4}$ & $\mathbf{2.17} \cdot \mathbf{10}^{-4}$ & $\mathbf{6.85} \cdot \mathbf{10}^{-5}$ & $\mathbf{3.95} \cdot \mathbf{10}^{-5}$ \\ \bottomrule
\end{tabular}
\caption{Mean squared error of UNet-style models on the shallow water equations experiment.}
\label{tab:mse_sw_unet}
\end{table}

\begin{figure}
    \centering
    \begin{subfigure}[b]{\columnwidth}
    \includegraphics[width=\textwidth]{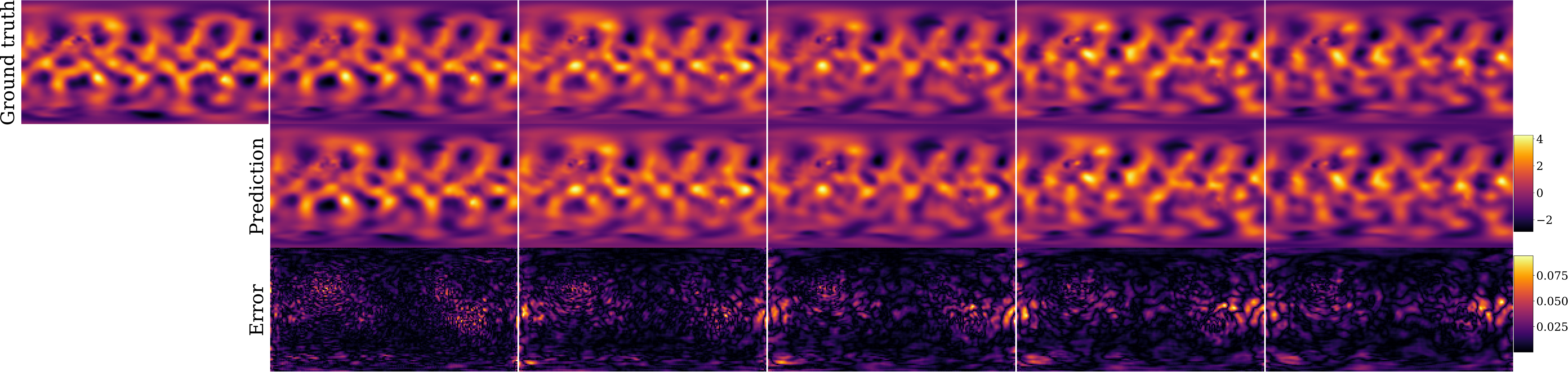}
    \caption{Pressure}
    \end{subfigure}
    \begin{subfigure}[b]{\columnwidth}
    \includegraphics[width=\textwidth]{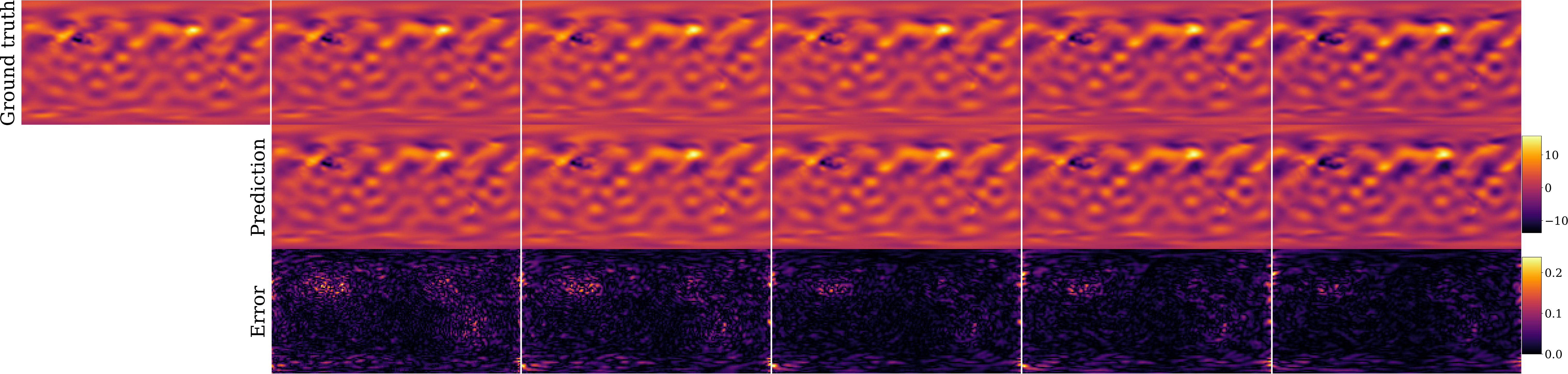}
    \caption{Eastward wind speed}
    \end{subfigure}
    \begin{subfigure}[b]{\columnwidth}
    \includegraphics[width=\textwidth]{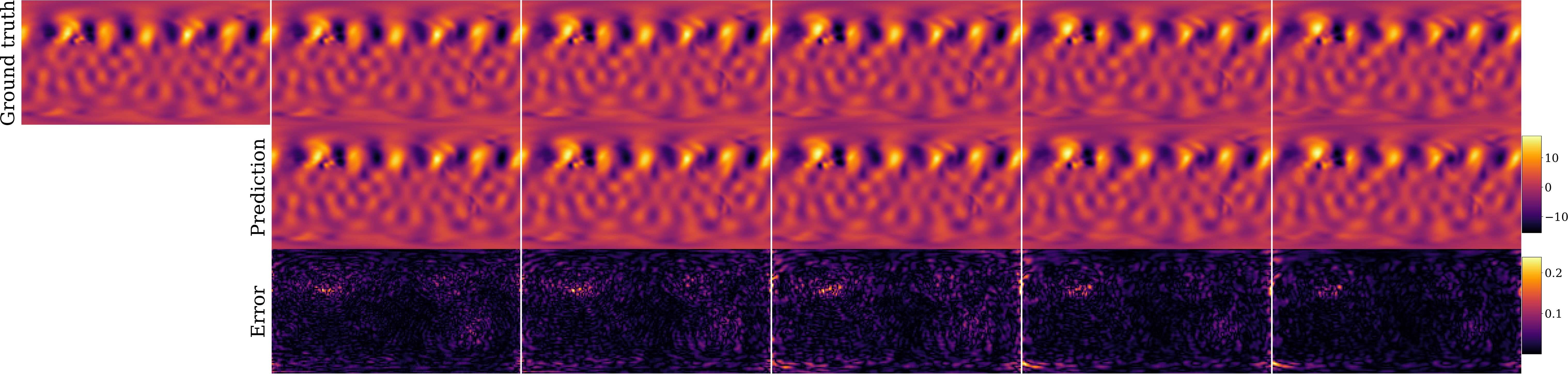}
    \caption{Northward wind speed}
    \end{subfigure}
    \caption{Example rollout from GCA-UNet on the 6 hour shallow water equations.}
    \label{fig:example_shallowwater_gcaunet}
\end{figure}

\begin{figure}
    \centering
    \begin{subfigure}[b]{\columnwidth}
    \includegraphics[width=\textwidth]{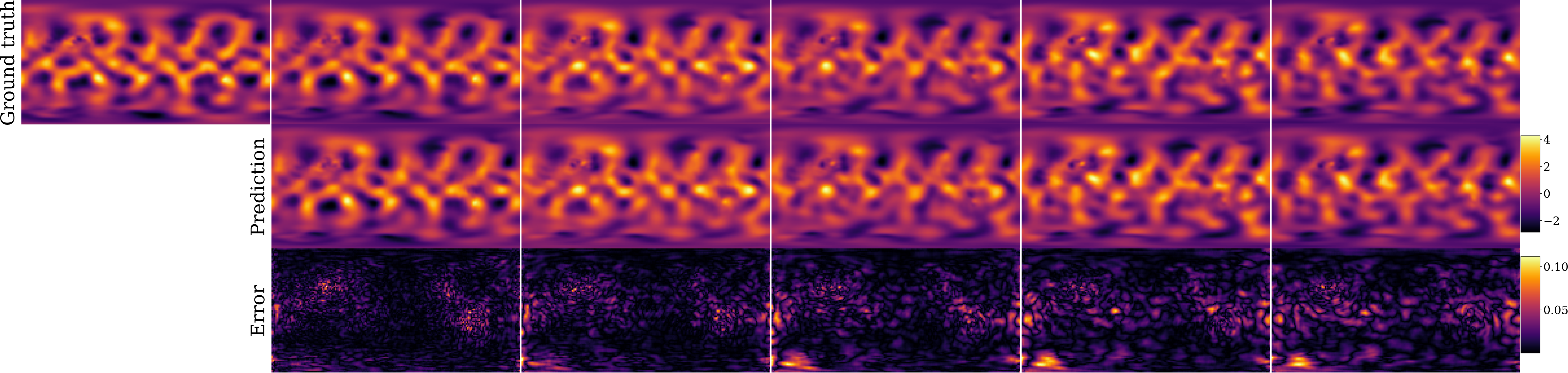}
    \caption{Pressure}
    \end{subfigure}
    \begin{subfigure}[b]{\columnwidth}
    \includegraphics[width=\textwidth]{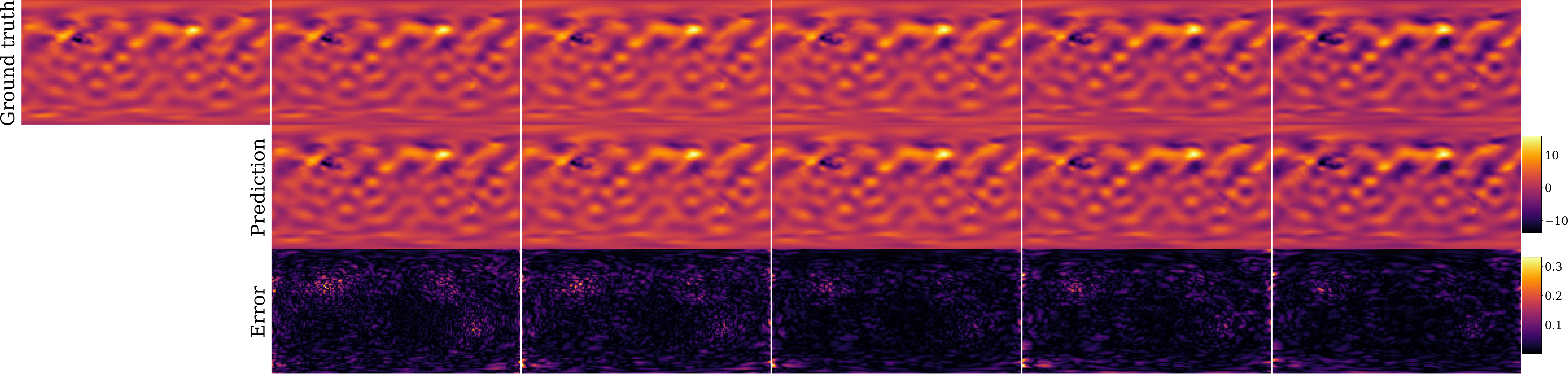}
    \caption{Eastward wind speed}
    \end{subfigure}
    \begin{subfigure}[b]{\columnwidth}
    \includegraphics[width=\textwidth]{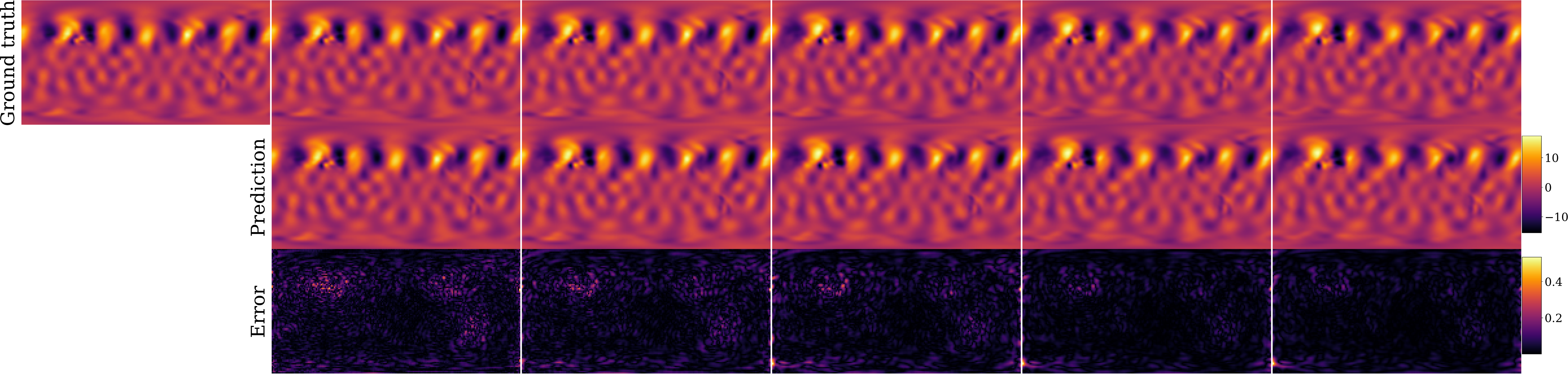}
    \caption{Northward wind speed}
    \end{subfigure}
    \caption{Example rollout from UNet on the 6 hour shallow water equations. Note that the bottom left corner of the pressure field visibly differs from our GCA-UNet rollout in \Cref{fig:example_shallowwater_gcaunet}.}
    \label{fig:example_shallowwater_unet}
\end{figure}

Finally, we show an examplary predicted and ground-truth trajectory for our GCA-UNet in \Cref{fig:example_shallowwater_gcaunet} and for the baseline UNet in \Cref{fig:example_shallowwater_unet}.

\textbf{Runtimes. }

\begin{table}[H]
\footnotesize
\centering
\begin{tabular}{@{}llllll@{}}
\toprule
                       & F (s/it) & B (s/it) & F + B (s/it) & Parameters (M) & Compute                  \\ \midrule
ResNet                 & 0.10           & 0.17            & 0.27                       & 3              & $2 \times 4$ NVIDIA V100 \\
CResNet                & 0.63           & 0.95            & 1.59                       & 3              & $2 \times 4$ NVIDIA V100 \\
CResNet$_{\text{rot}}$ & 0.62           & 0.94            & 1.56                       & 3              & $2 \times 4$ NVIDIA V100 \\
GCA-ResNet             & 0.13           & 0.19            & 0.33                       & 3              & $2 \times 4$ NVIDIA V100 \\
UNet                   & 0.25           & 0.40            & 0.65                       & 58             & $2 \times 4$ NVIDIA V100 \\
CUNet                  & 0.64           & 0.97            & 1.61                       & 58             & $2 \times 4$ NVIDIA V100 \\
CUNet$_{\text{rot}}$   & 0.63           & 0.95            & 1.59                       & 58             & $2 \times 4$ NVIDIA V100 \\
GCA-UNet               & 0.42           & 0.61            & 1.03                       & 58             & $2 \times 4$ NVIDIA V100 \\ \bottomrule
\end{tabular}
\caption{Overview of the estimated forward and backward runtimes of the implemented models.}
\label{tab:runtimes}
\end{table}

\subsection{Navier-Stokes}
The incompressible Navier-Stokes equations are built upon momentum and mass conservation of fluids. For the velocity flow field $v$, the incompressible Navier-Stokes equations read
\begin{align}
    \frac{\partial v}{\partial t} &= -v \cdot \nabla v + \mu \nabla^2 v - \nabla p + f \ , \\
    \nabla \cdot v &= 0 \ ,
    \label{eq:Navier_Stokes}
\end{align}
where $v \cdot \nabla v$ is the convection of the fluid, $\mu \nabla^2 v$ the diffusion controlled via the viscosity parameter $\nu$ of the fluid, $\nabla p$ the internal pressure, and $f$ an external buoyancy force. Convection is the rate of change of a vector field along a vector field (in this case along itself), and diffusion is the net movement form higher valued regions to lower concentration regions.
Additional to the velocity field $v(x)$, we introduce a scalar field $s(x)$ 
representing a scalar quantity, such as particle concentration or smoke density in our case, that is being \emph{advected}, i.e., transported along the velocity field.  

We implement the 2D Navier-Stokes equation using \texttt{${\Phi}$Flow}\footnote{\url{https://github.com/tum-pbs/PhiFlow}} \citep{holl2020phiflow}. Solutions are obtained by solving for the pressure field, and subsequently subtracting the gradients of the pressure field. 

We obtain data on a closed domain with Dirichlet boundary conditions ($v=0$) for the vector (velocity), and Neumann boundaries $\frac{\partial s}{\partial x} = 0$ for the scalar field.
The grid has a spatial resolution of $128 \times 128$ ($\Delta x=0.25$, $\Delta y=0.25$), and temporal resolution of $\Delta t = 1.5s$. The viscosity is set to $\nu=0.01$. The scalar field is initialized with random Gaussian noise fluctuations, and the velocity field is initialized to $0$. 
We run the simulation for $21.0s$ and sample every $1.5s$. Trajectories contain scalar smoke density, and vector velocity fields at $14$ different time points. 

\textbf{GCAN implementation. }
We use the algebra $\sG_{3, 0, 0}$ and therefore use the same implementation as for the shallow water experiment.

\textbf{Objective. }
Similar to the shallow water equations, we have
\begin{align}
    \mathcal{L}_{\text{MSE}} := \frac{1}{N_y} \sum_{t=1}^{N_t} \sum_{n=1}^{N_{\text{fields}}} \sum_{y \in \mathbb{Y}} (x_{tny} - \hat{x}_{tny})^2.
\end{align}

\textbf{Model selection and optimization. }
We used the best-performing models in the shallow water equations: the GCA-UNet and UNet and compare them at different number of training trajectories: 832, 2080, and 5200.
For the UNet, we tested a version with 64 input channels and 72 input channels, which matches the parameter of the GCA counterpart (58M).
We searched across learning rates of $2 \cdot 10^{-4}$ and $5 \cdot 10^{-4}$, or all architectures including the baseline UNets.
For the GCA-UNet, we further tested normalization schemes and activation schemes, obtaining similar results to the shallow water experiment.

We used the Adam optimizer with best performing learning rate, and cosine annealing \citep{loshchilov2016sgdr} with linear warmup. We trained at all data-regimes for 50 epochs.

\textbf{Extended results. }
We present in \Cref{tab:navier_stokes} the numerical results of our Navier-Stokes experiment also presented in the main paper.
\begin{table}[H]
\centering
\begin{tabular}{@{}llll@{}}
\toprule
Training trajectories    & 832     & 2080     & 5200    \\ \midrule
UNet            & 0.00290 & 0.001040 & $9.71\cdot 10^{-4}$ \\
GCA-UNet (Ours) & \textbf{0.00194} & $\mathbf{9.48\cdot 10^{-4}}$ & $\mathbf{9.21\cdot 10^{-4}}$ \\ \bottomrule
\end{tabular}
\caption{Mean squared error of UNet models on the Navier-Stokes experiment.}
\label{tab:navier_stokes}
\end{table}

\section{Clifford Algebra}
\label{sec:supp_clifford}

This section provides a pedagogical insight into how a Clifford algebra and its multiplication rules are constructed in a basis-independent way.
Consider a vector space $V$ whose characteristic is not $2$.
A Clifford algebra over $V$ has an distributive and associative bilinear product with the construction
\begin{align}
v^2 - \langle v, v \rangle 1=0,
\label{eq:cl_crit}
\end{align}
for $v \in V$.
Here, $\langle \cdot, \cdot \rangle$ denotes a symmetric quadratic form, and $1$ is the multiplicative identity.
In other words, multiplication of a vector with itself identifies with the quadratic form.
This yields the identity
\begin{align}
    (u+v)^2 = \langle u+v, u+v \rangle,
\end{align}
from which the fundamental Clifford identity (geometric product) arises:
\begin{align}
    uv = 2\langle u, v \rangle - vu.
\end{align}
As such, we see that 
\begin{align}
    \frac12 (uv + vu) = \langle u, v \rangle.
\end{align}
Further, we can construct an antisymmetric part 
\begin{align}
    u \wedge v := \frac12 (uv - vu), \label{eq:wedge_def}
\end{align}
with 
\begin{align}
u \wedge v = -v \wedge u.
\end{align}
This is commonly referred to as \emph{wedge product}.
As such, we see that we get the identity
\begin{align}
    uv = \langle u, v \rangle + u \wedge v,
\end{align}
which is referred to as the \emph{geometric product} and directly follows from the fundamental Clifford identity.
In \Cref{supp:ga} we show how the geometric product is computed in practice after choosing a basis.

\section{Pseudocode}

\newcommand{\chin}{c_\text{in}}
\newcommand{\chout}{c_\text{out}}
\begin{algorithm}[ht]
\caption{Pseudocode for obtaining Clifford kernels to compute geometric products. Here, $\mv a \in \sG_{p, q, r}^{\chout \times \chin}$, where $\chin$ is the number of input and $\chout$ the number of output channels of the layer. A Cayley table $M\in\{-1, 0, 1\}^{n \times n \times n}$ ($n=p+q+r$) is provided by the algebra and its signature to effectively compute geometric products.
}
\SetKwProg{Fn}{Function}{:}{}
\SetKwFunction{FKernel}{GetCliffordKernel}
\Fn{\FKernel{$M$: Cayley Table, $\mv a$: Action}}{
   $K^{\text{left}}_{jvkw} \gets \sum_i  M_{ijk}\mv a_{vwi}$  \; \\
   $K^{\text{right}}_{jviw} \gets \sum_k M_{ijk}\mv a_{vwk}$  \; \\
    \KwRet $K^{\text{left}}$, $K^{\text{right}}$\; 
}
\end{algorithm}

\begin{algorithm}[ht]
\caption{Pseudocode for the geometric algebra conjugate linear layer.}
\SetKwProg{Fn}{Function}{:}{}
\SetKwFunction{FConjugateLinear}{ConjugateLinear}
\Fn{\FConjugateLinear{$x$: Input, $\mv a$: Actions, $w$: Weights, $A$: GeometricAlgebra}}{
    $M \gets$ \texttt{CayleyTable}$(A)$ \; \\
    $\mv a^{-1} \gets$ \texttt{Reverse}$(\mv a)$ \; \\
    $\mv x \gets $ \texttt{Embed}$(A, x)$ \; \\
    $K^{\text{left}}, \_ \gets \texttt{GetCliffordKernel}(M, \mv a)$ \; \\
    $\_, K^{\text{right}} \gets \texttt{GetCliffordKernel}(M, \mv a^{-1})$ \; \\
    $\mv x \gets $ \texttt{GroupActionLinear}$(w, K^{\text{left}}, K^{\text{right}}, \mv x)$ \; \\
    $x \gets \texttt{Retrieve}(A, \mv x)$ \; \\
    \KwRet $x$ \;
}
\end{algorithm}

\begin{algorithm}[!ht]
\caption{Pseudocode for the multivector sigmoid linear units.}
\SetKwProg{Fn}{Function}{:}{}
\SetKwFunction{FMSiLU}{MSiLU}
\Fn{\FMSiLU{$x$: Input, $w$: Weights, $A$: GeometricAlgebra, Agg: Aggregation}}{
    $\mv x \gets $ \texttt{Embed}$(A, x)$ \; \\
    \For{$[\mv x]_k$ in $\mv x$}{
        \If{Agg $=$ ``linear''}{$[\mv x]_k \gets \sigma\left(\texttt{Linear}(\mv x, w_k)\right) \cdot [\mv x]_k$} \; 
        \If{Agg $=$ ``sum''}{$[\mv x]_k \gets \sigma   \left(\texttt{Sum}(\mv x) \right) \cdot [\mv x]_k$}\; 
        \If{Agg $=$ ``mean''}{$[\mv x]_k \gets \sigma \left(\texttt{Mean}(\mv x) \right) \cdot [\mv x]_k$} \;
    }
    $x \gets \texttt{Retrieve}(A, \mv x)$ \; \\
    \KwRet $x$ \;
}
\end{algorithm}

\begin{algorithm}[ht]
\caption{Pseudocode for the geometric algebra normalization layer.}
\SetKwProg{Fn}{Function}{:}{}
\SetKwFunction{FGANormalize}{GCANormalize}
\Fn{\FGANormalize{$x$: Input, $s$: Rescaling, $A$: GeometricAlgebra}}{
    $\mv x \gets $ \texttt{Embed}$(A, x)$ \; \\
    \For{$[\mv x]_k$ in $\mv x$}{
        $[\mv x]_k \gets s_k \cdot \left([\mv x]_k - \texttt{ChannelsAverage}([\mv x]_k])\right) / \, \texttt{ChannelsAverage}(\lVert [\mv x]_k \rVert)$
    }
    $x \gets \texttt{Retrieve}(A, \mv x)$ \; \\
    \KwRet $x$ \;
}
\end{algorithm}

\begin{algorithm}[ht]
\caption{Pseudocode for the $\sG_{3, 0, 0}$ conjugate linear layer.}
\SetKwProg{Fn}{Function}{:}{}
\SetKwFunction{FG300ConjugateLinear}{$\sG_{3, 0, 0}$ conjugate linear layer.}
\Fn{\FConjugateLinear{$x$: Input, $\mv a$: Actions, $w$: Weights, $A$: GeometricAlgebra}}{
    $\mv x \gets $ \texttt{Embed}$(A, x)$ \; \\
    $\b R \gets \texttt{GetRotationalKernel}(\mv a)$ \; \\
    $\mv x \gets $ \texttt{RotationalConvolution}$(w, \b R, \mv x)$ \; \\
    $x \gets \texttt{Retrieve}(A, \mv x)$ \; \\
    \KwRet $x$ \;
}
\end{algorithm}

\end{document}